\pdfoutput=1
\RequirePackage{fix-cm}
\documentclass[10pt,twocolumn,letterpaper]{article}

\usepackage[pagenumbers]{cvpr}

\usepackage{amssymb,booktabs,colortbl,xcolor,pifont,multirow}
\usepackage[T1]{fontenc}
\usepackage{bm}
\usepackage{graphicx}
\usepackage{subcaption}
\usepackage{tabularx,array}
\usepackage{enumitem}
\usepackage{overpic}
\usepackage{needspace}

\definecolor{SphereTint}{HTML}{EAF2F2}
\newcolumntype{C}{>{\centering\arraybackslash}X}

\definecolor{cvprblue}{rgb}{0.21,0.49,0.74}
\usepackage[pagebackref,breaklinks,colorlinks,allcolors=cvprblue]{hyperref}
\hypersetup{
    pdftitle={Enhancing Medium-range Global Weather Forecasting with Spherical Harmonic Test-Time Calibration},
    pdfauthor={Ziyu Zhou, Jianyu Lai, Tian Zhou, Yuxuan Liang, James Kwok}
}

\title{Enhancing Medium-range Global Weather Forecasting with Spherical Harmonic Test-Time Calibration}

\author{
Ziyu Zhou$^{1}$ \quad
Jianyu Lai$^{1}$ \quad
Tian Zhou$^{2}$ \quad
Yuxuan Liang$^{3}$ \quad
James Kwok$^{1*}$\\[2pt]
$^{1}$The Hong Kong University of Science and Technology \quad
$^{2}$Ant Group\\
$^{3}$The Hong Kong University of Science and Technology (Guangzhou)\\[2pt]
{\tt\small $^{*}$Corresponding author \quad zzhou651@connect.ust.hk \quad jamesk@cse.ust.hk}
}

\begin{document}
\maketitle
\begin{abstract}
Accurate medium-range global weather forecasting is critical for disaster preparedness and resource management, yet deployed data-driven forecasters still exhibit systematic errors that vary across variables, regions, and lead times. Although the ground truth of issued forecasts becomes available continuously after deployment, existing neural calibrators either fix their parameters after offline training or rely on iterative gradient updates that trade off adaptation speed, stability, and computational cost. In this paper, we propose \textbf{SphereTTC}, the first gradient-free test-time calibration framework that refines the outputs of frozen global weather forecasters within a truncated spherical harmonic subspace using delayed ground truth. It couples: (1) a spherical harmonic residual correction with smooth spectral tapering, which constrains the spatial complexity of each adjustment while preserving fine-scale structure; (2) a closed-form, identity-regularized affine update estimated from a rolling history of forecast--ground-truth pairs; and (3) a chronological validation gate that sets the correction strength by evaluating corrections fitted on earlier pairs against more recent ones. Extensive experiments on ERA5 show that SphereTTC consistently reduces the overall nRMSE of all 11 forecasting backbones by 4.2\% on average and outperforms existing calibration methods in error reduction.
\end{abstract}

\section{Introduction}
\label{sec:introduction}
\vspace{-1mm}

\begin{figure}[t]
    \centering
    \includegraphics[
        width=\columnwidth,
        trim=15 10 13
        10,
        clip
    ]{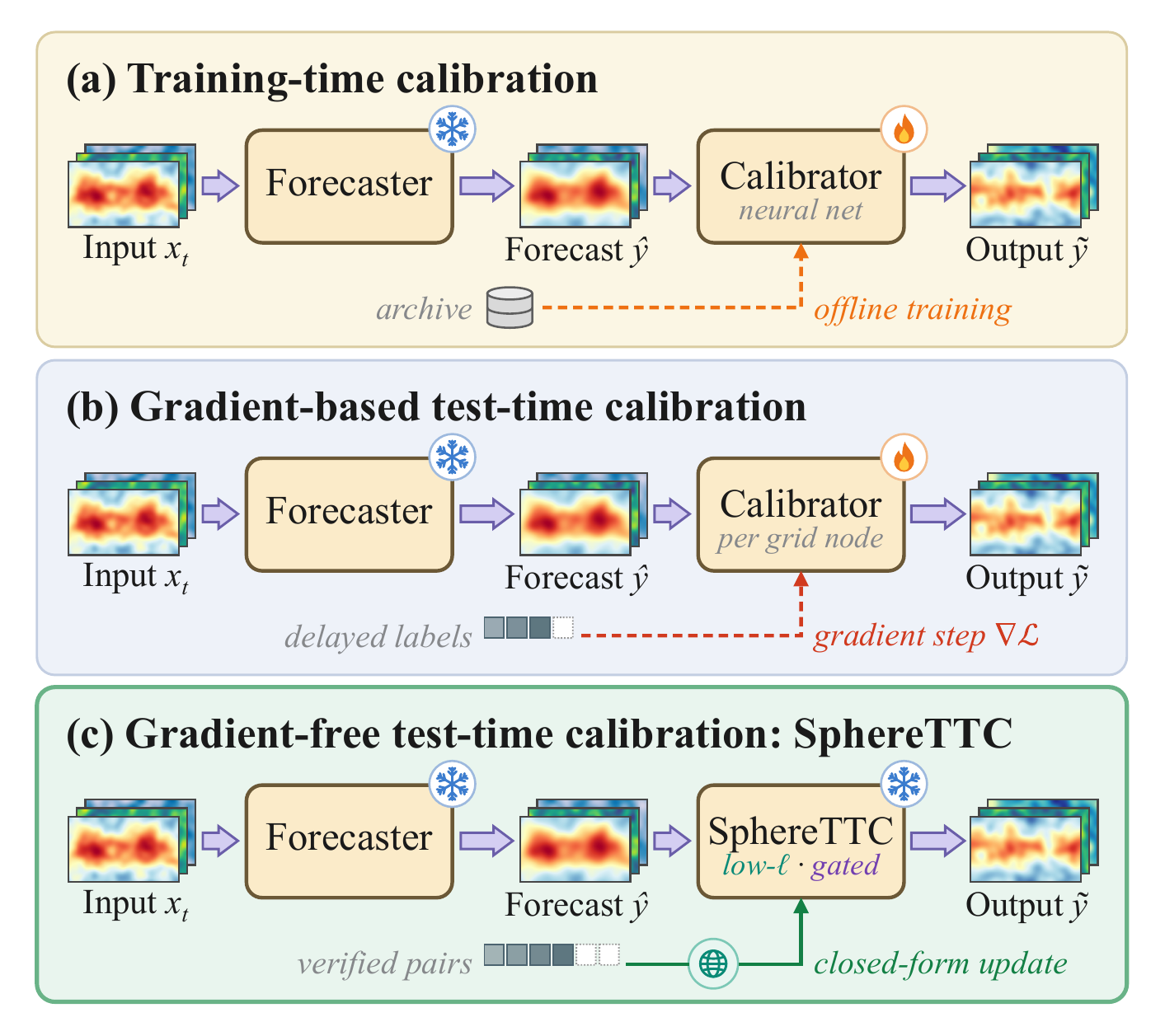}
    \caption{Calibration paradigms for frozen forecasters. Specifically, (a) Training-time calibration fixes an offline-trained network, (b) Gradient-based test-time calibration updates a network online and (c) SphereTTC fits spherical harmonic corrections in closed form and gates them by chronological validation.}
    \label{fig:calibration_paradigms}
\end{figure}

Medium-range global weather forecasting predicts the atmospheric state from one day to about two weeks ahead~\cite{bauer2015quiet,chen2023fengwu}.
Data-driven forecasters built on Fourier neural operators~\cite{pathak2022fourcastnet,bonev2023sfno}, transformers~\cite{STORM,lang2024aifs,chen2026emformer}, and global--regional coupling~\cite{chen2026stcast} now match or surpass operational numerical weather prediction at a fraction of its cost~\cite{bi2023accurate,lam2023learning,han2024fengwughr}.
Despite this progress, deployed forecasters still exhibit systematic errors that vary across variables, regions, and lead times.
Meanwhile, the ground truth of issued forecasts arrives continuously after deployment, providing direct feedback for correcting subsequent forecasts.

To exploit this feedback, post-deployment calibration estimates corrections from historical forecasts and their verified ground truth~\cite{glahn1972mos,vannitsem2021statistical,mouatadid2023adaptive}.
Recent neural calibrators follow two main paradigms.
Training-time calibration trains an auxiliary network offline and keeps it fixed after deployment (Figure~\ref{fig:calibration_paradigms}(a)), as in PnP-Corrector~\cite{pnpcorrector} and ASTAFN~\cite{xu2026astafn}.
Gradient-based test-time calibration instead updates its modules online as new verification arrives (Figure~\ref{fig:calibration_paradigms}(b)), as in TAFAS~\cite{kim2025tafas} and ST-TTC~\cite{chen2025stttc}.

However, calibrating with delayed ground truth remains challenging.
First, offline networks cannot adapt to error patterns that emerge after deployment, while online gradient updates must balance adaptation speed, stability, and cost.
Second, the ground truth of a forecast arrives only after its lead time has elapsed, so corrections that reduced past errors may not help later forecasts, especially when unconstrained or grid-point-wise updates overfit short-lived local errors.
Hence, corrections should have bounded spatial complexity, and their benefit should be verified before they are applied.
Spherical harmonics provide the former as an orthonormal basis on the sphere ordered by spatial scale~\cite{bonev2023sfno}, and the sequential arrival of ground truth enables the latter through chronological validation on recent history.

To alleviate these problems, we propose \textbf{SphereTTC}, a model-agnostic test-time calibration framework for frozen global weather forecasters (Figure~\ref{fig:calibration_paradigms}(c)).
For each variable and lead time, SphereTTC maintains a \textbf{verified memory} of past forecast--ground-truth pairs.
A \textbf{spherical harmonic correction} then estimates identity-regularized affine corrections in closed form within a truncated spherical harmonic subspace, requiring neither neural network training nor backpropagation.
Smooth spectral tapering attenuates adjustments near the truncation boundary, and the residual formulation keeps fine-scale structures outside the subspace intact.
Finally, a \textbf{chronological validation gate} fits candidate corrections on earlier pairs and evaluates them on more recent ones, suppressing adjustments whose historical gains do not persist.
On ERA5, SphereTTC reduces the forecast error of all 11 backbones and achieves larger error reductions than existing calibration methods.

Our main contributions are as follows:

\begin{itemize}[leftmargin=*]
    \item We identify that existing neural calibrators neither regulate the spatial complexity of their corrections nor verify their transfer to subsequent forecasts, both of which are critical when calibrating with delayed ground truth.

    \item We propose SphereTTC, the first gradient-free test-time calibration method that corrects forecasters within a truncated spherical harmonic subspace. Unlike gradient-based calibrators, it updates all parameters in closed form and gates each correction by chronological validation.

    \item Extensive experiments on ERA5 show that SphereTTC achieves a 4.20\% mean overall nRMSE reduction across 11 backbones with no neural parameters and outperforms other spatio-temporal calibration methods.
\end{itemize}

\section{Related Work}
\label{sec:related_work}
\vspace{-1mm}

\noindent\textbf{Data-driven Weather Forecasting.}
Deep learning has been widely applied to Earth observation and atmospheric data, ranging from remote sensing change detection~\cite{chen2020spatial,chen2022remote,chen2022adversarial} and climate data compression~\cite{han2025climate} to global weather forecasting~\cite{pathak2022fourcastnet,lang2024aifs,chen2023fengwu}.
Modern data-driven weather models encode the spatial structure of the atmosphere through graph meshes, Earth-specific attention, and spectral operators.
For example, GraphCast~\cite{lam2023learning} exchanges information on a multiscale mesh, and Pangu~\cite{bi2023accurate} combines three-dimensional Earth-specific attention with hierarchical temporal aggregation.
FuXi~\cite{chen2023fuxi} cascades models over successive forecast ranges, and SFNO~\cite{bonev2023sfno} formulates global convolutions with spherical harmonics to respect the geometry of the Earth.
Recent models further reach kilometer-scale resolution~\cite{han2024fengwughr}, incorporate new variables incrementally~\cite{chen2025vamoe}, and improve long-term skill~\cite{chen2026emformer}.
These approaches improve the forecaster itself, whereas SphereTTC is complementary: it uses spherical harmonics as a restricted correction space for a frozen forecaster rather than learning a new dynamical model.

\noindent\textbf{Forecast Correction and Test-time Adaptation.}
Statistical post-processing relates past forecast errors to subsequent corrections, ranging from model output statistics~\cite{glahn1972mos} to adaptive bias correction~\cite{mouatadid2023adaptive}.
Neural correction increases the expressiveness of this mapping; for example, PnP-Corrector~\cite{pnpcorrector} trains an external correction agent around frozen simulators to mitigate error amplification in coupled forecasting.
More broadly, test-time and cross-domain adaptation address distribution shifts between training and deployment~\cite{sun2020ttt,wang2021tent,han2020focus}.
In forecasting, TAFAS~\cite{kim2025tafas} exploits partially observed ground truth with gated calibration modules, and ST-TTC~\cite{chen2025stttc} adapts temporal-frequency amplitude and phase through streaming gradient updates.
In contrast, SphereTTC solves a bounded spherical harmonic correction in closed form and uses a chronological holdout to decide how strongly to apply it to the next forecast.
\section{Preliminaries}
\label{sec:preliminary}
\vspace{-1mm}

\begin{figure*}[t]
    \centering
    \includegraphics[width=\textwidth]{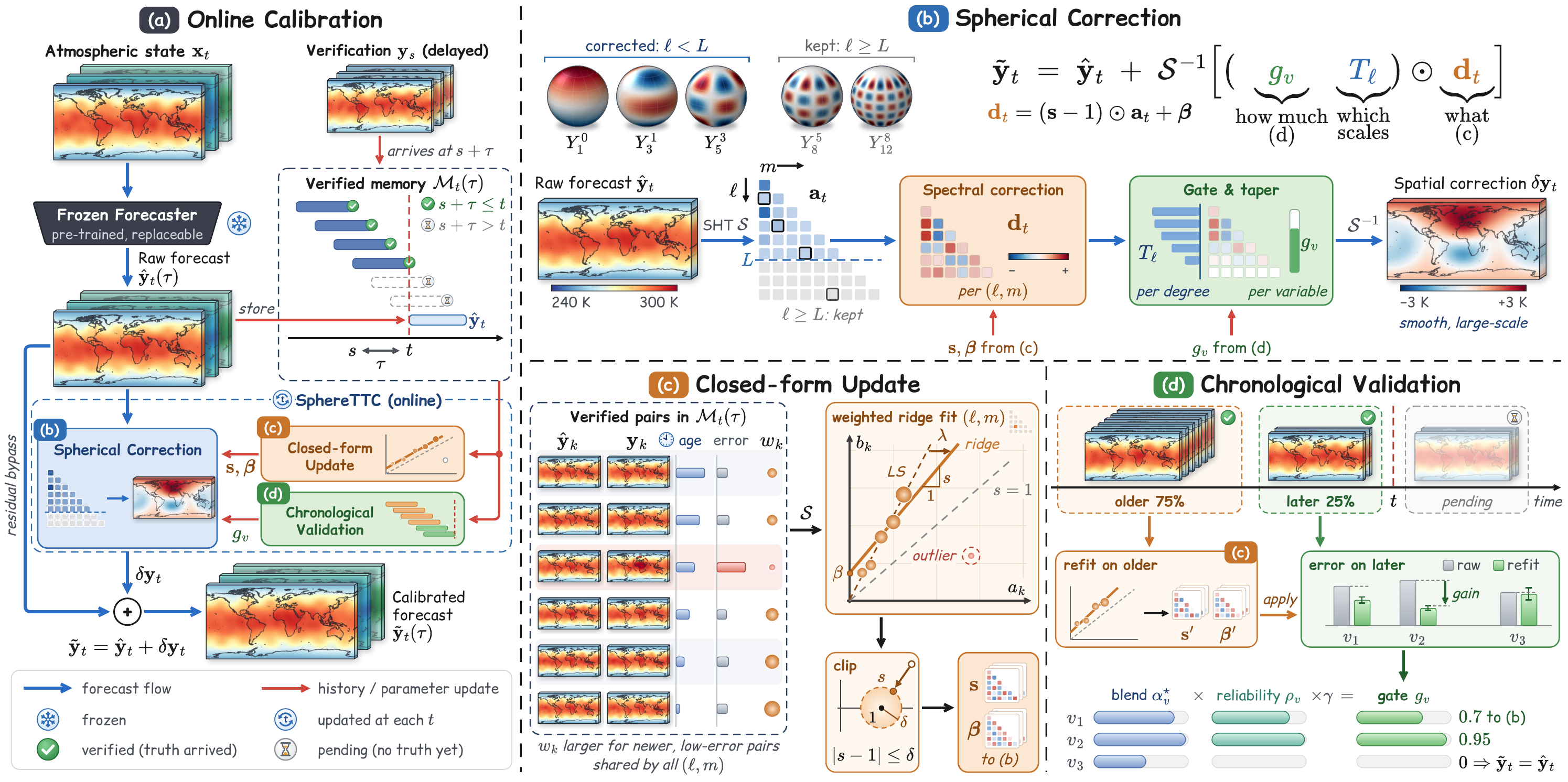}
    \caption{Main framework of SphereTTC. (a) Verified memory of forecast--ground-truth pairs. (b) Residual correction in a truncated spherical harmonic subspace. (c) Closed-form affine update under an identity-centered constraint. (d) Chronological validation gating of the correction strength, using earlier pairs for fitting and later pairs for validation.}
    \label{fig:spherettc}
\end{figure*}

\subsection{Problem Definition}
\label{sec:problem}
\vspace{-1mm}

We consider global weather forecasting of $V$ meteorological variables on a latitude--longitude grid with $H$ latitude bands and $W$ longitude points, where each variable at a distinct pressure level is treated as a separate channel.
We use bold symbols for arrays and regular type for their scalar entries.
Let $\mathbf{X}_t\in\mathbb{R}^{V\times H\times W}$ denote the atmospheric state at time $t$.
A pre-trained and frozen forecaster $\mathcal{F}$ produces a forecast at lead time $\tau$ as:
\begin{equation}
    \widehat{\mathbf{Y}}_t(\tau)
    =\mathcal{F}(\mathbf{X}_{\leq t};\tau)
    \in\mathbb{R}^{V\times H\times W},
    \label{eq:forecast}
\end{equation}
where $\mathbf{X}_{\leq t}$ denotes the observation history required by the forecaster.
The corresponding verification target is denoted by $\mathbf{Y}_t(\tau)$, indexed by the forecast issuance time $t$ and the lead time $\tau$.
This notation accommodates both direct and autoregressive forecasting backbones~\cite{bi2023accurate,lam2023learning}.

Test-time calibration uses newly available verification to improve subsequent forecasts.
Let $r_s(\tau)$ be the time at which the target $\mathbf{Y}_s(\tau)$ becomes available.
For instantaneous verification without an additional reporting delay, $r_s(\tau)=s+\tau$. For a daily mean indexed by target-day start $t_{\mathrm{valid},s}(\tau)$, $r_s(\tau)=t_{\mathrm{valid},s}(\tau)+24\,\mathrm{h}+d$, where $d\geq0$ is the reporting latency. At issuance time $t$, a calibrator may only use historical pairs with $r_s(\tau)\leq t$, and its goal is to produce a corrected forecast $\widetilde{\mathbf{Y}}_t(\tau)$ that is more accurate with respect to the subsequently observed target.
The forecasting backbone remains fixed throughout deployment, while the calibrator is updated from this verified history. The same availability rule applies to gate validation and warmup.

\subsection{Spherical Harmonic Representation}
\label{sec:spherical_harmonics}
\vspace{-1mm}

Spherical harmonics provide a spectral representation of fields on the sphere~\cite{driscoll1994computing,nistDLMF}.
For a square-integrable scalar field $f:\mathbb{S}^2\rightarrow\mathbb{R}$, the functions $\{Y_\ell^m\}_{\ell\geq0,\,-\ell\leq m\leq\ell}$ form a complete orthonormal basis~\cite{nistDLMF,simons2006spatiospectral}.
For a point $\bm{\omega}\in\mathbb{S}^2$ with colatitude $\theta$ and longitude $\varphi$, the transform and reconstruction are given by~\cite{driscoll1994computing}:
\begin{equation}
\begin{aligned}
    a_{\ell m}
    &=\int_{\mathbb{S}^2}
      f(\bm{\omega})\overline{Y_\ell^m(\bm{\omega})}\,\mathrm{d}\Omega,\\
    f(\bm{\omega})
    &=\sum_{\ell=0}^{\infty}\sum_{m=-\ell}^{\ell}
      a_{\ell m}Y_\ell^m(\bm{\omega}),
\end{aligned}
\label{eq:sht}
\end{equation}
where $\mathrm{d}\Omega=\sin\theta\,\mathrm{d}\theta\,\mathrm{d}\varphi$ and the overline denotes complex conjugation; the reconstruction holds in the $L^2$ sense.
The degree $\ell$ determines the spatial scale: small degrees describe broad structures, whereas large degrees describe fine variations~\cite{simons2006spatiospectral,bonev2023sfno}.
The order $m$ determines the longitudinal structure, with $|m|$ oscillations around a latitude circle~\cite{nistDLMF}.
This scale ordering follows from the eigenvalue relation $-\Delta_{\mathbb{S}^2}Y_\ell^m=\ell(\ell+1)Y_\ell^m$~\cite{simons2006spatiospectral}.

For real fields, $a_{\ell,-m}=(-1)^m\overline{a_{\ell m}}$, so only the nonnegative orders need to be stored~\cite{nistDLMF}.
We denote by $\mathcal{S}_L$ the transform that retains degrees $0\leq\ell<L$, and by $\mathcal{S}_L^{-1}$ the associated real-valued synthesis~\cite{simons2006spatiospectral,mcewen2011sampling}.
The coefficients are stored in an $L\times L$ array with valid entries $0\leq m\leq\ell<L$ and zeros elsewhere.
On the discrete grid, the transform uses spherical quadrature~\cite{driscoll1994computing,mcewen2011sampling,bonev2023sfno}.
\section{Methodology}
\label{sec:methodology}
\vspace{-1mm}

Figure~\ref{fig:spherettc} illustrates the framework of SphereTTC, which comprises four components: (i) a \textbf{Verified Memory} of forecast--ground-truth pairs whose targets are available, (ii) a \textbf{Spherical Correction} restricted to a truncated spherical harmonic subspace, (iii) a \textbf{Closed-Form Update} of the affine parameters, and (iv) a \textbf{Chronological Validation Gate} that determines the correction strength.

\subsection{Online Calibration with Verified Memory}
\label{sec:memory}
\vspace{-1mm}

For each lead time $\tau$, we retain the latest $K$ eligible forecast--verification pairs in chronological order:
\begin{equation}
    \mathcal{M}_t(\tau)
    =\bigl\{\bigl(\widehat{\mathbf{Y}}_{s_k}(\tau),
                       \mathbf{Y}_{s_k}(\tau)\bigr)\bigr\}_{k=1}^{n},
    \label{eq:memory}
\end{equation}
where $n\leq K$, $s_1<\cdots<s_n<t$, and $r_{s_k}(\tau)\leq t$.
Each stored prediction is the raw output issued by the frozen backbone, and a pending forecast enters the verified memory only when its target becomes available.
Under a common verification delay, longer lead times therefore rely on older forecast issuances.
Figure~\ref{fig:spherettc} illustrates the timing for instantaneous targets, for which $r_s(\tau)=s+\tau$; daily-mean verification follows the availability rule in the Appendix.

At each issuance, SphereTTC updates the calibration parameters and the gate from $\mathcal{M}_t(\tau)$ and applies them to the current raw forecast.
Calibration is performed separately for each backbone, lead time, and variable.
In the following, we fix a lead time and omit $\tau$, as well as the dependence of the fitted parameters and gates on $t$.
Historical coefficient pairs are cached as $\mathbf{A}_k=\mathcal{S}_L(\widehat{\mathbf{Y}}_{s_k})$ and $\mathbf{B}_k=\mathcal{S}_L(\mathbf{Y}_{s_k})$.
For variable $v$, $\mathbf{A}_{k,v}\in\mathbb{C}^{L\times L}$ denotes its coefficient matrix with scalar entries $a_{k,v,\ell m}$, and analogous notation applies to $\mathbf{B}_{k,v}$.
Coefficient norms and averages below are taken over the stored nonnegative-order array.

\subsection{Spherical Correction with Scale Control}
\label{sec:correction}
\vspace{-1mm}

A global forecast contains structures at many spatial scales, whereas a short verified history provides limited evidence for estimating their corrections.
We therefore parameterize the correction by spatial scale using spherical harmonics and restrict its support to degrees $\ell<L$.
This makes the calibration complexity explicit while preserving the remaining forecast information through a residual connection.

Formally, let $\mathbf{A}_t=\mathcal{S}_L(\widehat{\mathbf{Y}}_t)$ be the coefficients of the current forecast, and for variable $v$, let $\mathbf{S}_v,\bm{\beta}_v\in\mathbb{C}^{L\times L}$ collect the per-mode scales and offsets.
The candidate correction is defined as:
\begin{equation}
    \mathbf{D}_{t,v}
    =(\mathbf{S}_v-\mathbf{1})\odot\mathbf{A}_{t,v}+\bm{\beta}_v,
    \label{eq:direction}
\end{equation}
where $\mathbf{1}$ is the all-ones matrix and $\odot$ denotes element-wise multiplication.
Its scalar entries satisfy $d_{t,v,\ell m}=(s_{v,\ell m}-1)a_{t,v,\ell m}+\beta_{v,\ell m}$.
We attenuate this correction toward the truncation boundary using a fixed degree-dependent taper:
\begin{equation}
    T_\ell=
    \begin{cases}
        \displaystyle\cos^p\!\left(\frac{\pi\ell}{2(L-1)}\right),
            &0\leq\ell<L,\\
        0, &\ell\geq L,
    \end{cases}
    \label{eq:taper}
\end{equation}
where $L\geq2$ and $p>0$ controls the attenuation profile.
The taper matrix $\mathbf{T}\in\mathbb{R}^{L\times L}$ has entries $T_{\ell m}=T_\ell$ for $m\leq\ell$ and zero elsewhere, so all valid orders at the same degree share one attenuation factor.
A variable-specific gate $g_v$ then determines the calibrated output:
\begin{equation}
    \widetilde{\mathbf{Y}}_{t,v}
    =\widehat{\mathbf{Y}}_{t,v}
      +\mathcal{S}_L^{-1}\!\left(g_v\mathbf{T}\odot\mathbf{D}_{t,v}\right).
\label{eq:output}
\end{equation}
Equivalently, the applied coefficient correction is $\delta a_{t,v,\ell m}=g_vT_\ell d_{t,v,\ell m}$.
Since only this correction is synthesized from the truncated coefficients, the original forecast supplies all spatial information outside the correction subspace.
Here, the affine parameters determine \emph{what} to change, the taper determines the degree-wise attenuation, and the gate determines \emph{how much} correction to apply.

\subsection{Constrained Closed-Form Update}
\label{sec:closed_form}
\vspace{-1mm}

We estimate the affine parameters from the verified memory using weighted ridge regression~\cite{hoerl1970ridge} with an identity-centered penalty.
The weights emphasize recent pairs while limiting the influence of pairs with unusually large forecast errors.
Specifically, for one variable (suppressing $v$), we define the residual energy $e_k=L^{-2}\|\mathbf{B}_k-\mathbf{A}_k\|_F^2$ and assign:
\begin{equation}
    w_k=2^{-(n-k)/h}
      \min\!\left\{1,\frac{\kappa\,\operatorname{median}_j e_j}
                              {\max(e_k,\varepsilon)}\right\},
\label{eq:weights}
\end{equation}
where $h>0$ is a half-life measured in memory entries, $\kappa>0$ controls the outlier attenuation, and $\varepsilon>0$ is a numerical floor that is also applied to the median.
The weights are normalized as $\bar{w}_k=w_k/\sum_j w_j$.
One weight per pair and variable is shared across all spherical harmonic coefficients.

For each valid mode $(\ell,m)$, we then solve:
\begin{equation}
    \min_{\substack{s,\beta\in\mathbb{C}\\|s-1|\leq\delta}}
    \sum_{k=1}^{n}\bar{w}_k
        |s a_{k,\ell m}+\beta-b_{k,\ell m}|^2
    +\lambda|s-1|^2,
    \label{eq:ridge_objective}
\end{equation}
where $\lambda>0$ pulls the multiplicative scale toward one and $\delta>0$ bounds its deviation from the identity.
The additive offset is estimated jointly.
Given the weights and the regularization strength, these fits are separable across modes, and no cross-mode coefficient matrix is estimated.

Let $\mu^a_{\ell m}=\sum_k\bar{w}_k a_{k,\ell m}$ and $\mu^b_{\ell m}=\sum_k\bar{w}_k b_{k,\ell m}$ be the weighted means, and define the centered moments:
\begin{equation}
\begin{aligned}
    q_{\ell m}
    &=\sum_k\bar{w}_k|a_{k,\ell m}-\mu^a_{\ell m}|^2,\\
    c_{\ell m}
    &=\sum_k\bar{w}_k
      (b_{k,\ell m}-\mu^b_{\ell m})
      \overline{(a_{k,\ell m}-\mu^a_{\ell m})}.
\end{aligned}
\label{eq:moments}
\end{equation}
The regularization strength is scaled to the coefficient variance as $\lambda=\lambda_0 L^{-2}\sum_{\ell,m}q_{\ell m}$, with a positive floor on the mean variance and a dimensionless hyperparameter $\lambda_0>0$.
The constrained solution is:
\begin{equation}
\begin{aligned}
    s^{\circ}_{\ell m}
    &=\frac{c_{\ell m}+\lambda}{q_{\ell m}+\lambda},\qquad
      s_{\ell m}=\Pi_\delta(s^{\circ}_{\ell m}),\\
    \beta_{\ell m}&=\mu^b_{\ell m}-s_{\ell m}\mu^a_{\ell m},
\end{aligned}
\label{eq:ridge_solution}
\end{equation}
where $\Pi_\delta$ is the Euclidean projection onto the complex disk $\{z:|z-1|\leq\delta\}$.
In other words, the unconstrained ridge solution is clipped around one, and the offset is computed using the clipped scale.
These statistics and projections yield a closed-form update without gradient optimization.

\subsection{Chronological Validation Gating}
\label{sec:gating}
\vspace{-1mm}

A correction that fits the available history may not transfer to subsequent forecasts.
We assess this transfer by splitting verified memory chronologically into an older fitting subset $\mathcal{I}_{\mathrm{fit}}$ and a later validation subset $\mathcal{I}_{\mathrm{val}}$. The nominal shares are 75\% and 25\%, subject to a minimum validation size.
Applying Section~\ref{sec:closed_form} to the older subset yields temporary parameters $(\mathbf{S}'_v,\bm{\beta}'_v)$.
For each later pair, the raw error is $\mathbf{E}_{k,v}=\mathbf{A}_{k,v}-\mathbf{B}_{k,v}$, and the candidate correction is $\mathbf{D}_{k,v}=(\mathbf{S}'_v-\mathbf{1})\odot\mathbf{A}_{k,v}+\bm{\beta}'_v$.
The gate is estimated from this affine candidate before tapering, and all sums over $k$ below are restricted to $\mathcal{I}_{\mathrm{val}}$. Estimating the gate prior to tapering keeps validation aligned with Eq.~\eqref{eq:ridge_objective} and exposes near-boundary modes directly to reliability screening, while $\mathbf{T}$ adds conservative boundary smoothing at inference.

\noindent\textbf{Optimal Blend.}
We first find the scalar blend that minimizes the total squared coefficient error over the later pairs:
\begin{equation}
    \alpha_v^\star
    =\left[
      -\frac{\sum_k
          \operatorname{Re}\langle\mathbf{E}_{k,v},\mathbf{D}_{k,v}\rangle_F}
        {\sum_k
          \|\mathbf{D}_{k,v}\|_F^2}
     \right]_0^1,
    \label{eq:blend}
\end{equation}
where $[u]_0^1=\min(1,\max(0,u))$ and $\langle\cdot,\cdot\rangle_F$ is the Hermitian Frobenius inner product.
A zero correction direction is assigned a zero blend.

\noindent\textbf{Reliability Weighting.}
We also evaluate how consistently the full affine candidate improves on the raw forecast.
Its gain on validation pair $k$ is:
\begin{equation}
    \Delta_{k,v}
    =\frac{\|\mathbf{E}_{k,v}\|_F^2
           -\|\mathbf{E}_{k,v}+\mathbf{D}_{k,v}\|_F^2}{L^2}.
    \label{eq:gain}
\end{equation}
Let $\mu_{\Delta,v}$ and $\sigma_{\Delta,v}$ be the mean and population standard deviation of these gains, and let $N_{\mathrm{val}}=|\mathcal{I}_{\mathrm{val}}|$.
Writing the mean raw error as $\bar{r}_v=(N_{\mathrm{val}}L^2)^{-1}\sum_k\|\mathbf{E}_{k,v}\|_F^2$, we set:
\begin{equation}
    \rho_v
    =\left[
       \frac{\mu_{\Delta,v}
             -z\,\sigma_{\Delta,v}/\sqrt{N_{\mathrm{val}}}}
            {|\mu_{\Delta,v}|+0.01\bar{r}_v}
      \right]_0^1,\quad g_v=\gamma\,\alpha_v^\star\rho_v,
\label{eq:gate}
\end{equation}
where $z\geq0$ penalizes the variability of validation gains and $\gamma\in[0,1]$ sets the maximum correction strength.
Numerical denominators use small positive floors.
As a result, weak or inconsistent historical gains reduce the applied correction, and a nonpositive variability-adjusted gain sets $g_v=0$.
The gate is shared by all retained modes of a variable and is recomputed at each issuance.

\noindent\textbf{Final Update and Complexity.}
After estimating the gate, we refit $(\mathbf{S},\bm{\beta})$ on all pairs in the current verified memory and apply Eq.~\eqref{eq:output}.
All available verification therefore contributes to the final parameter estimate, while the validation gains are computed from the earlier-only fit.
When too few verified pairs are available for fitting or validation, the gate is set to zero.
With cached coefficients, fitting and gating require $O(KVL^2)$ time, and the verified memory occupies $O(KVL^2)$ space per lead time, in addition to the forward and inverse spherical harmonic transforms.
Operating entirely on forecast outputs, SphereTTC requires neither backbone gradients nor an offline-trained auxiliary network.

\section{Experiments}
\label{sec:experiments}
\vspace{-1mm}

\begin{table*}[t]
\centering
\captionsetup{skip=4pt}
\begingroup
\definecolor{SphereRawBest}{RGB}{36,82,161}
\definecolor{SphereCalBest}{RGB}{181,75,0}
\definecolor{SphereWithout}{RGB}{95,102,112}
\definecolor{SphereDelta}{RGB}{239,239,239}
\newcommand{\SphereNo}{\textcolor{SphereWithout}{\ding{55}}}
\newcommand{\SphereYes}{\textcolor{SphereCalBest}{\ding{51}}}
\newcommand{\SphereChange}[2]{%
  \makebox[3.0em][c]{\makebox[0.6em][c]{$#1$}%
  \hspace{0.15em}\makebox[2.25em][c]{\textbf{#2}}}}
\caption{Raw (\SphereNo) and calibrated (\SphereYes) scores. Shading shows relative changes. \textcolor{SphereRawBest}{\textbf{Blue}} and \textcolor{SphereCalBest}{\textbf{orange}} mark the respective best scores.}
\label{tab:main_results}
\fontsize{7.5}{9.2}\selectfont
\setlength{\tabcolsep}{0.85pt}
\renewcommand{\arraystretch}{1.18}
\begin{tabular*}{\textwidth}{@{\extracolsep{0pt}}c@{\hspace{4pt}}c@{\hspace{6pt}}*{5}{c@{\hspace{2pt}\extracolsep{0pt}}c@{\hspace{3.5pt}}>{\columncolor{SphereDelta}[0.7pt][0.7pt]}c@{\hspace{4pt}\extracolsep{\fill}}}c@{\hspace{2pt}\extracolsep{0pt}}c@{\hspace{3.5pt}}>{\columncolor{SphereDelta}[0.7pt][0.7pt]}c@{}}
\toprule
\multicolumn{2}{c}{Models} & \multicolumn{3}{c}{\textbf{ConvLSTM}~\cite{shi2015convolutional}} & \multicolumn{3}{c}{\textbf{Transformer}~\cite{vaswani2017attention}} & \multicolumn{3}{c}{\textbf{ClimODE}~\cite{verma2024climode}} & \multicolumn{3}{c}{\textbf{FourCastNetV2}~\cite{bonev2023sfno}} & \multicolumn{3}{c}{\textbf{FuXi}~\cite{chen2023fuxi}} & \multicolumn{3}{c}{\textbf{GraphCast}~\cite{lam2023learning}} \\
\cmidrule(lr){3-5}\cmidrule(lr){6-8}\cmidrule(lr){9-11}\cmidrule(lr){12-14}\cmidrule(lr){15-17}\cmidrule(lr){18-20}
\multicolumn{2}{c}{w/ SphereTTC} & \SphereNo & \SphereYes & \cellcolor{white}$\Delta$(\%) & \SphereNo & \SphereYes & \cellcolor{white}$\Delta$(\%) & \SphereNo & \SphereYes & \cellcolor{white}$\Delta$(\%) & \SphereNo & \SphereYes & \cellcolor{white}$\Delta$(\%) & \SphereNo & \SphereYes & \cellcolor{white}$\Delta$(\%) & \SphereNo & \SphereYes & \cellcolor{white}$\Delta$(\%) \\
\midrule
\multirow{2}{*}{Day 1} & nRMSE & 0.3346 & 0.3226 & \SphereChange{\downarrow}{3.59} & 0.3318 & 0.3179 & \SphereChange{\downarrow}{4.19} & 0.3403 & 0.3218 & \SphereChange{\downarrow}{5.43} & 0.2665 & 0.1841 & \SphereChange{\downarrow}{30.91} & 0.1500 & 0.1309 & \SphereChange{\downarrow}{12.70} & \textcolor{SphereRawBest}{\textbf{0.1133}} & \textcolor{SphereCalBest}{\textbf{0.1095}} & \SphereChange{\downarrow}{3.35} \\
 & ACC & 0.7430 & 0.7652 & \SphereChange{\uparrow}{2.98} & 0.7550 & 0.7782 & \SphereChange{\uparrow}{3.07} & 0.7708 & 0.7897 & \SphereChange{\uparrow}{2.46} & 0.8898 & 0.9111 & \SphereChange{\uparrow}{2.40} & 0.9425 & 0.9482 & \SphereChange{\uparrow}{0.61} & \textcolor{SphereRawBest}{\textbf{0.9728}} & \textcolor{SphereCalBest}{\textbf{0.9740}} & \SphereChange{\uparrow}{0.13} \\
\midrule
\multirow{2}{*}{Day 3} & nRMSE & 0.4781 & 0.4740 & \SphereChange{\downarrow}{0.86} & 0.4682 & 0.4637 & \SphereChange{\downarrow}{0.97} & 0.5013 & 0.4927 & \SphereChange{\downarrow}{1.70} & 0.3479 & 0.2641 & \SphereChange{\downarrow}{24.10} & 0.2323 & 0.2046 & \SphereChange{\downarrow}{11.91} & \textcolor{SphereRawBest}{\textbf{0.1788}} & \textcolor{SphereCalBest}{\textbf{0.1778}} & \SphereChange{\downarrow}{0.58} \\
 & ACC & 0.3962 & 0.4102 & \SphereChange{\uparrow}{3.54} & 0.4366 & 0.4504 & \SphereChange{\uparrow}{3.16} & 0.3890 & 0.3989 & \SphereChange{\uparrow}{2.53} & 0.8454 & 0.8595 & \SphereChange{\uparrow}{1.68} & 0.9020 & 0.9095 & \SphereChange{\uparrow}{0.83} & \textcolor{SphereRawBest}{\textbf{0.9352}} & \textcolor{SphereCalBest}{\textbf{0.9359}} & \SphereChange{\uparrow}{0.07} \\
\midrule
\multirow{2}{*}{Day 5} & nRMSE & 0.5036 & 0.5013 & \SphereChange{\downarrow}{0.46} & 0.4972 & 0.4950 & \SphereChange{\downarrow}{0.46} & 0.5058 & 0.5040 & \SphereChange{\downarrow}{0.35} & 0.4474 & 0.3663 & \SphereChange{\downarrow}{18.14} & 0.3236 & 0.2922 & \SphereChange{\downarrow}{9.71} & \textcolor{SphereRawBest}{\textbf{0.2706}} & \textcolor{SphereCalBest}{\textbf{0.2680}} & \SphereChange{\downarrow}{0.95} \\
 & ACC & 0.2710 & 0.2795 & \SphereChange{\uparrow}{3.12} & 0.3028 & 0.3094 & \SphereChange{\uparrow}{2.18} & 0.2855 & 0.2890 & \SphereChange{\uparrow}{1.22} & 0.7365 & 0.7461 & \SphereChange{\uparrow}{1.30} & 0.8324 & 0.8386 & \SphereChange{\uparrow}{0.75} & \textcolor{SphereRawBest}{\textbf{0.8595}} & \textcolor{SphereCalBest}{\textbf{0.8607}} & \SphereChange{\uparrow}{0.13} \\
\midrule
\multirow{2}{*}{Day 7} & nRMSE & 0.5128 & 0.5107 & \SphereChange{\downarrow}{0.41} & 0.5072 & 0.5052 & \SphereChange{\downarrow}{0.39} & 0.5277 & 0.5256 & \SphereChange{\downarrow}{0.40} & 0.5427 & 0.4536 & \SphereChange{\downarrow}{16.41} & 0.4108 & 0.3737 & \SphereChange{\downarrow}{9.05} & \textcolor{SphereRawBest}{\textbf{0.3785}} & \textcolor{SphereCalBest}{\textbf{0.3646}} & \SphereChange{\downarrow}{3.68} \\
 & ACC & 0.2158 & 0.2239 & \SphereChange{\uparrow}{3.76} & 0.2453 & 0.2508 & \SphereChange{\uparrow}{2.25} & 0.2152 & 0.2190 & \SphereChange{\uparrow}{1.78} & 0.5884 & 0.5921 & \SphereChange{\uparrow}{0.62} & 0.7291 & \textcolor{SphereCalBest}{\textbf{0.7339}} & \SphereChange{\uparrow}{0.66} & \textcolor{SphereRawBest}{\textbf{0.7293}} & 0.7321 & \SphereChange{\uparrow}{0.39} \\
\midrule
\multirow{2}{*}{Day 10} & nRMSE & 0.5202 & 0.5182 & \SphereChange{\downarrow}{0.37} & 0.5147 & 0.5130 & \SphereChange{\downarrow}{0.32} & 0.5242 & 0.5229 & \SphereChange{\downarrow}{0.26} & 0.6455 & 0.5360 & \SphereChange{\downarrow}{16.96} & \textcolor{SphereRawBest}{\textbf{0.5100}} & \textcolor{SphereCalBest}{\textbf{0.4660}} & \SphereChange{\downarrow}{8.64} & 0.5173 & 0.4720 & \SphereChange{\downarrow}{8.76} \\
 & ACC & 0.1741 & 0.1821 & \SphereChange{\uparrow}{4.61} & 0.2008 & 0.2052 & \SphereChange{\uparrow}{2.20} & 0.1788 & 0.1828 & \SphereChange{\uparrow}{2.20} & 0.3809 & 0.3772 & \SphereChange{\downarrow}{0.98} & \textcolor{SphereRawBest}{\textbf{0.5501}} & \textcolor{SphereCalBest}{\textbf{0.5523}} & \SphereChange{\uparrow}{0.40} & 0.4973 & 0.4992 & \SphereChange{\uparrow}{0.37} \\
\midrule
\multirow{2}{*}{Overall} & nRMSE & 0.4698 & 0.4653 & \SphereChange{\downarrow}{0.96} & 0.4638 & 0.4589 & \SphereChange{\downarrow}{1.05} & 0.4798 & 0.4734 & \SphereChange{\downarrow}{1.34} & 0.4500 & 0.3608 & \SphereChange{\downarrow}{19.82} & 0.3253 & 0.2935 & \SphereChange{\downarrow}{9.80} & \textcolor{SphereRawBest}{\textbf{0.2917}} & \textcolor{SphereCalBest}{\textbf{0.2784}} & \SphereChange{\downarrow}{4.57} \\
 & ACC & 0.3600 & 0.3722 & \SphereChange{\uparrow}{3.38} & 0.3881 & 0.3988 & \SphereChange{\uparrow}{2.76} & 0.3679 & 0.3759 & \SphereChange{\uparrow}{2.18} & 0.6882 & 0.6972 & \SphereChange{\uparrow}{1.31} & 0.7912 & 0.7965 & \SphereChange{\uparrow}{0.67} & \textcolor{SphereRawBest}{\textbf{0.7988}} & \textcolor{SphereCalBest}{\textbf{0.8004}} & \SphereChange{\uparrow}{0.19} \\
\bottomrule
\end{tabular*}
\endgroup
\end{table*}

\begin{figure*}[t]
    \centering
    \captionsetup{skip=3pt}
    \includegraphics[width=\textwidth]{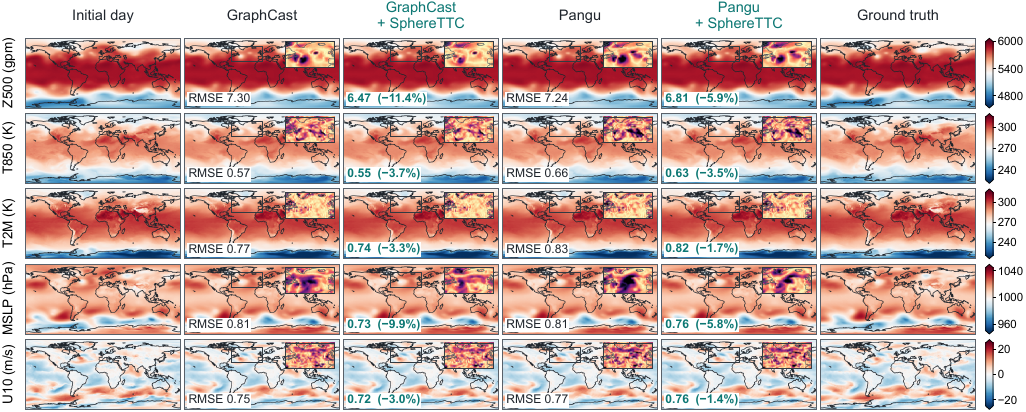}

    \caption{One-day forecast fields of GraphCast~\cite{lam2023learning} and Pangu~\cite{bi2023accurate} initialized on 30 May 2018, before and after SphereTTC. Each row shares a color scale; labels report the RMSE and its reduction, and insets show absolute errors (darker is larger).}
    \label{fig:spatial_forecast}
\end{figure*}

\subsection{Experimental Setup}
\label{sec:experimental_setup}
\vspace{-1mm}

\noindent\textbf{Dataset.}
We evaluate SphereTTC on the ERA5 reanalysis dataset~\cite{hersbach2020era5} at $1.5^{\circ}$ resolution, resulting in a $121\times240$ latitude--longitude grid.
The 49 variables comprise geopotential (Z), specific humidity (Q), temperature (T), and winds (U and V) at 50, 100, 200, 300, 500, 700, 850, 925, and 1000\,hPa, plus 2-m temperature (T2M), 10-m winds (U10 and V10), and mean sea-level pressure (MSLP).
The evaluation covers 730 daily initializations in 2018--2019 at leads of 1, 3, 5, 7, and 10 days.

\noindent\textbf{Forecasting Backbones.}
We apply SphereTTC to 11 forecasting backbones of two categories:
(1) \textbf{general-purpose architectures}: ConvLSTM~\cite{shi2015convolutional}, Transformer~\cite{vaswani2017attention}, FNO~\cite{FNO}, and ViT~\cite{dosovitskiy2020image};
(2) \textbf{weather-specific architectures}: ClimODE~\cite{verma2024climode}, CirT~\cite{Cirt}, FourCastNetV2~\cite{bonev2023sfno}, OneForecast~\cite{gao2025oneforecast}, FuXi~\cite{chen2023fuxi}, Pangu~\cite{bi2023accurate}, and GraphCast~\cite{lam2023learning}.
Table~\ref{tab:main_results} reports six representative backbones, and the results for all 11 are provided in the Appendix.

\noindent\textbf{Metrics.}
Following WeatherBench~\cite{rasp2020weatherbench,rasp2024weatherbench2}, we adopt the latitude-weighted Root Mean Square Error (RMSE) and Anomaly Correlation Coefficient (ACC).
For each variable and lead time, weighted errors and anomaly statistics are pooled over test initializations, with anomalies defined relative to the 1979--2016 calendar-day climatology.
Each RMSE is normalized by the standard deviation of its channel over the same period (nRMSE), and the Overall score averages all variables equally at each lead time and then across the five lead times.
As detailed in the Appendix, relative changes are computed from unrounded scores.

\noindent\textbf{Implementation Details.}
The six local backbones train from scratch on four NVIDIA H100 GPUs with 1979--2016 training data and 2017 validation. The five released models use official weights, including the $1^{\circ}$ GraphCast-small checkpoint.
All backbone weights remain frozen and SphereTTC selects hyperparameters solely on 2017 verification (though Pangu and OneForecast's published validation splits overlap 2018--2019). For availability, archived 49-channel results use target-date admission, whereas matched replays wait for target-day completion. The Appendix details both rules.

\noindent\textbf{Analysis Settings.}
For further analyses, we select five variables covering the upper-air and surface targets of WeatherBench~\cite{rasp2020weatherbench,rasp2024weatherbench2}: Z500 and T850 characterize synoptic-scale circulation and lower-tropospheric temperature, while T2M, MSLP, and U10 represent near-surface temperature, pressure, and wind.
Matched forecasts from GraphCast and Pangu in WeatherBench~2~\cite{rasp2024weatherbench2} let us test whether the same field-level gains recur across distinct backbone architectures.
Unless otherwise stated, we analyze 287 common dates in 2018 at lead times of 1, 3, 5, and 7 days.

\subsection{Overall Performance}
\label{sec:overall_performance}
\vspace{-1mm}

\begin{figure*}[t]
    \centering
    \captionsetup{skip=3pt}
    \includegraphics[width=\textwidth]{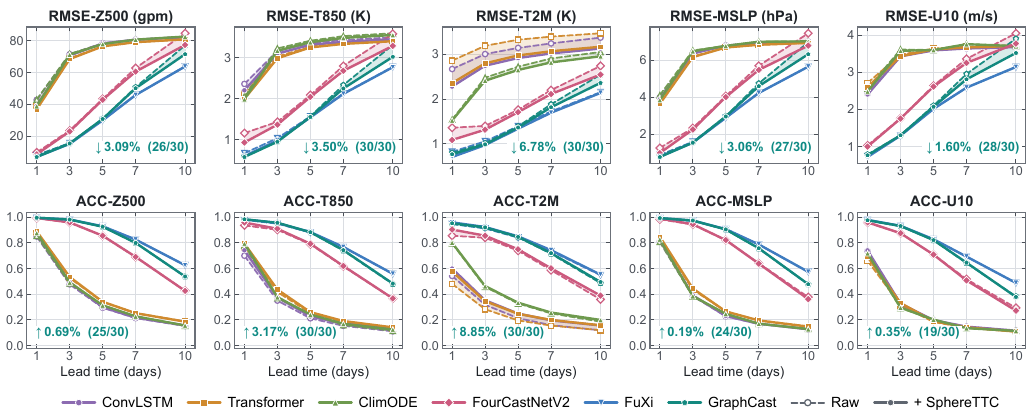}
    \caption{Forecast skill of six backbones across lead times. Dashed and solid curves show raw and calibrated forecasts, respectively; shading highlights their difference. Annotations report relative gains and counts of improved backbone--lead combinations.}
    \label{fig:lead_skill}
\end{figure*}

\begin{table}[t]
\centering
\captionsetup{skip=4pt}
\caption{Relative nRMSE and ACC gains (\%; higher is better).}
\label{tab:calibrator_comparison}
\fontsize{8.5}{10}\selectfont
\setlength{\tabcolsep}{2pt}
\renewcommand{\arraystretch}{1.00}
\begin{tabularx}{\columnwidth}{lCCCC}
\toprule
& \multicolumn{2}{c}{GraphCast~\cite{lam2023learning}} & \multicolumn{2}{c}{Pangu~\cite{bi2023accurate}}\\
\cmidrule(lr){2-3}\cmidrule(lr){4-5}
Method & nRMSE & ACC & nRMSE & ACC\\
& $\Delta$ (\%) & $\Delta$ (\%) & $\Delta$ (\%) & $\Delta$ (\%)\\
\midrule
\rowcolor{black!4}
\multicolumn{5}{l}{\textit{Training-time calibration}}\\
DRN~\cite{rasp2018neuralpostprocessing} & -0.87 & -0.041 & -0.12 & -0.117\\
BC-UNet~\cite{cui2025bcunet} & 0.24 & 0.032 & 0.65 & 0.087\\
PnP-Corrector~\cite{pnpcorrector} & 1.04 & 0.102 & 1.35 & 0.117\\
ASTAFN~\cite{xu2026astafn} & 1.10 & 0.107 & 0.55 & 0.092\\
\midrule
\rowcolor{black!4}
\multicolumn{5}{l}{\textit{Gradient-based test-time calibration}}\\
TAFAS~\cite{kim2025tafas} & 0.34 & 0.087 & 0.95 & 0.135\\
ST-TTC~\cite{chen2025stttc} & 0.21 & 0.079 & 0.99 & 0.142\\
COSA~\cite{im2026cosa} & 1.15 & 0.058 & 1.58 & 0.072\\
PETSA~\cite{medeiros2025petsa} & 1.11 & 0.115 & 0.48 & 0.238\\
\midrule
\rowcolor{black!4}
\multicolumn{5}{l}{\textit{Statistical post-processing}}\\
Running-mean bias & -0.68 & -0.033 & -0.14 & 0.127\\
Grid-point MOS~\cite{glahn1972mos} & -0.48 & -0.850 & 0.75 & -0.821\\
\midrule
\rowcolor{SphereTint}
\textbf{SphereTTC} & \textbf{1.19} & \textbf{0.125} & \textbf{1.75} & \textbf{0.173}\\
\bottomrule
\end{tabularx}
\end{table}

\begin{table}[t]
\centering
\captionsetup{skip=4pt}
\caption{Calibration-module costs. Params denotes neural parameters. Total includes update and application. Lower is better.}
\label{tab:calibration_cost}
\fontsize{8.5}{10}\selectfont
\setlength{\tabcolsep}{0.3pt}
\renewcommand{\arraystretch}{1.00}
\begin{tabularx}{\columnwidth}{lCCCC}
\toprule
Method & Params & Apply & Total & Peak\\
& (M) $\downarrow$ & (ms) $\downarrow$ & (ms) $\downarrow$ & (MB) $\downarrow$\\
\midrule
\rowcolor{black!4}
\multicolumn{5}{l}{\textit{Training-time calibration}}\\
DRN~\cite{rasp2018neuralpostprocessing} & 0.07 & 15.48 & 15.48 & 491.6\\
BC-UNet~\cite{cui2025bcunet} & 4.03 & 17.80 & 17.80 & 385.3\\
PnP-Corrector~\cite{pnpcorrector} & 141.07 & 1699.20 & 1699.20 & 2455.9\\
ASTAFN~\cite{xu2026astafn} & 124.65 & 6753.53 & 6753.53 & 3649.3\\
\midrule
\rowcolor{black!4}
\multicolumn{5}{l}{\textit{Gradient-based test-time calibration}}\\
TAFAS~\cite{kim2025tafas} & 0.73 & 4.44 & 16.80 & 192.7\\
ST-TTC~\cite{chen2025stttc} & 0.87 & 5.75 & 15.65 & 332.7\\
COSA~\cite{im2026cosa} & 0.41$\times 10^{-4}$ & 3.77 & 22.52 & 240.4\\
PETSA~\cite{medeiros2025petsa} & 10.02 & 6.58 & 27.62 & 223.5\\
\midrule
\rowcolor{SphereTint}
\textbf{SphereTTC} & \textbf{0} & \textbf{3.71} & \textbf{14.10} & \textbf{154.2}\\
\bottomrule
\end{tabularx}
\vspace{-2mm}
\end{table}

\noindent\textbf{Improvements across Backbones.}
SphereTTC reduces nRMSE on all six backbones in Table~\ref{tab:main_results}. Gains peak at day one for FourCastNetV2 and FuXi and at day ten for GraphCast.
Across all 49 channels, SphereTTC yields a \textbf{4.20\%} mean nRMSE reduction over 11 backbones (1.34\% median). Crucially, these gains are not confined to humidity: excluding it still improves every backbone (\textbf{2.06\%} mean, 0.97\% median; see the Appendix for FourCastNetV2 and FuXi humidity conversion before averaging).

\noindent\textbf{Field-level Comparison.}
Both one-day cases improve across all five fields (Figure~\ref{fig:spatial_forecast}). Z500 RMSE falls by 11.4\% on GraphCast and 5.9\% on Pangu. Gains are largest in Z500 and MSLP, consistent with corrections to broad circulation and pressure patterns. Improvements in T850, T2M, and U10 extend from the lower troposphere to the near surface for the same initializations. The Appendix complements these snapshots with 299 GraphCast dates, resolving where errors fall across latitude bands.

\noindent\textbf{Lead Times and Metrics.}
SphereTTC lowers nRMSE in \textbf{all 30} backbone--lead pairs and raises ACC in 29. The ACC ranking also changes at some leads. FuXi moves ahead of GraphCast at day seven. At day ten, FourCastNetV2 pairs a lower nRMSE with weaker anomaly agreement. Figure~\ref{fig:lead_skill} shows gains on both metrics for T850 and T2M in every pair. For U10, reductions in error magnitude occur more often than gains in anomaly agreement.

\subsection{Comparison with Calibration Methods}
\label{sec:calibrator_comparison}
\vspace{-1mm}

\noindent\textbf{Compared Methods.}
Table~\ref{tab:calibrator_comparison} compares training-time, test-time, and statistical calibrators on 238 common initializations (1 May--24 December 2018). Online tuning uses 41 validation origins through 22 April, with labels complete by 30 April. The Appendix details offline fitting, warm-up, and update schedules.

\noindent\textbf{Calibration Accuracy.}
In Table~\ref{tab:calibrator_comparison}, SphereTTC achieves the top nRMSE gains on both GraphCast (1.19\%) and Pangu (1.75\%) and the highest GraphCast ACC gain (0.125\%). In a separate matched Appendix comparison with archived COSA~\cite{im2026cosa} under four-lead updates, SphereTTC significantly improves both metrics at days~1 and~3 (95\% intervals excluding zero), yields higher day-5 ACC, and remains statistically comparable overall.

\subsection{Calibration Efficiency}
\label{sec:robustness_cost}
\vspace{-1mm}

\noindent\textbf{Parameters, Latency, and Memory.}
SphereTTC runs in 14.10\,ms on an L4 with zero neural parameters (Table~\ref{tab:calibration_cost}). Its 154.2\,MB peak is less than half that of any training-time baseline. At fixed bandwidth, transforms cost $O(HWL)$, with $O(KVL^2)$ fitting and gating on cached coefficients.

\begin{figure}[t]
    \centering
    \captionsetup{skip=7pt}
    \includegraphics[width=\columnwidth]{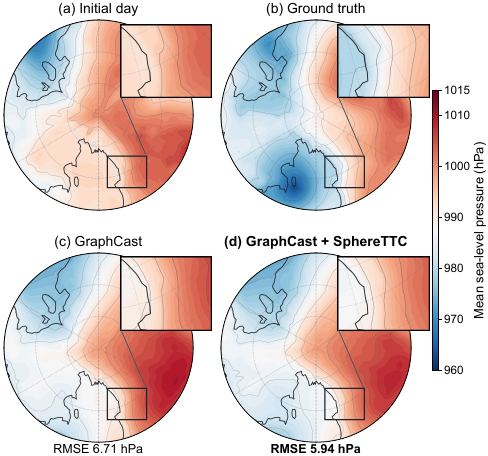}
    \caption{An illustrative seven-day GraphCast~\cite{lam2023learning} MSLP forecast over Antarctica before and after calibration. Matched insets magnify the same region, and all views share one pressure scale.}
    \label{fig:polar_comparison}
\end{figure}

\begin{figure}[t]
    \centering
    \captionsetup{skip=3pt}
    \includegraphics[width=\columnwidth]{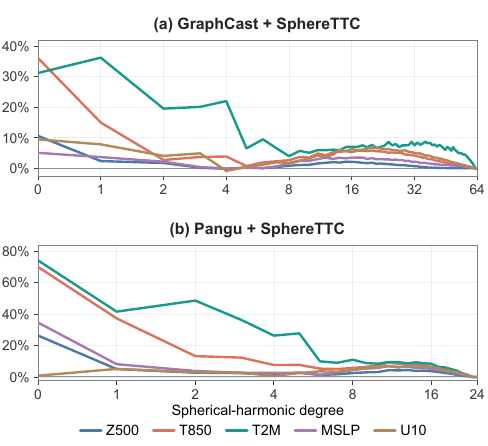}
    \caption{Error-energy reductions by scale vs. raw forecasts.}
    \label{fig:scale_diagnostics}
\end{figure}

\begin{figure}[t]
    \centering
    \captionsetup{skip=3pt}
    \includegraphics[width=\columnwidth]{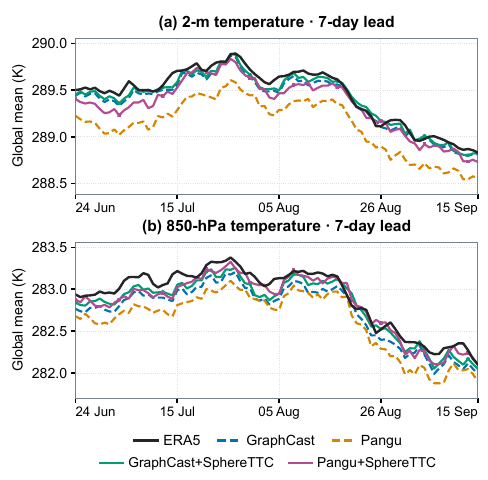}

    \captionsetup{skip=3.79pt}\caption{Seven-day temperature forecasts over an illustrative window of 84 consecutive verification dates. Solid and dashed model curves denote calibrated and raw forecasts, respectively.}
    \label{fig:temporal_variation}
\end{figure}

\subsection{Forecast Diagnostics}
\label{sec:diagnostics}
\vspace{-1mm}

\noindent\textbf{Polar Pressure Structure.}
Figure~\ref{fig:polar_comparison} follows a seven-day GraphCast MSLP forecast over Antarctica. The matched insets show SphereTTC weakening the positive pressure bias along the coast, where the verifying field develops lower pressure. Across the cap, mean bias falls from 4.20 to 3.28\,hPa and RMSE from 6.71 to 5.94\,hPa. The corrected coastal low is shallower than in the verifying field. More spatial diagnostics appear in the Appendix.

\noindent\textbf{Correction across Spatial Scales.}
SphereTTC enforces clean scale separation (Figure~\ref{fig:scale_diagnostics}): it cuts low-degree (0--4) T2M error energy by 23.7\% for GraphCast and 47.8\% for Pangu (improving all five fields), tapers smoothly toward the cutoff, and leaves finer scales virtually untouched.

\noindent\textbf{Temporal Variation.}
On unsmoothed seven-day forecasts (Figure~\ref{fig:temporal_variation}), SphereTTC removes persistent biases and closely tracks time-varying ERA5 means for both backbones, confirming agile online adaptation. Full-cohort day-5 GraphCast series RMSE also drops 16.7\% (T2M) and 8.5\% (T850) in the Appendix.

\noindent\textbf{Verification and Gating.}
In the Appendix, we additionally audit our holdout gate against four blend rules and a strict replay (which refits per historical issuance and scores tapered fields post-verification). Our gate yields the fewest degraded outcomes across rules, strict replay lowers nRMSE on both backbones, and GraphCast retains gains through 120\,h of extra delay.

\begin{table}[t]
\centering
\captionsetup{skip=4pt}
\caption{GraphCast ablation. Higher ($\uparrow$) is better for $\Delta\mathrm{nRMSE}$ and $\Delta\mathrm{ACC}$; lower ($\downarrow$) is better for degraded fraction (Degr.).}
\label{tab:component_analysis}
\fontsize{8.5}{10}\selectfont
\setlength{\tabcolsep}{2pt}
\renewcommand{\arraystretch}{1.00}
\begin{tabularx}{\columnwidth}{lCCCCC}
\toprule
& \multicolumn{3}{c}{$\Delta\mathrm{nRMSE}$ (\%) $\uparrow$} & $\Delta\mathrm{ACC}$ & Degr.\\
\cmidrule(lr){2-4}
Variant & Z500 & T850 & T2M & (\%) $\uparrow$ & (\%) $\downarrow$\\
\midrule
w/o gate & 0.17 & 1.29 & 1.57 & 0.13 & 25.59\\
w/o taper & 0.13 & 1.05 & 1.48 & 0.15 & 10.28\\
w/o identity ridge & 0.12 & 1.33 & 2.07 & 0.16 & 5.28\\
w/o recency & 0.13 & 1.21 & 1.47 & 0.14 & 2.24\\
w/o SH (grid-point) & 0.18 & 0.79 & 1.20 & 0.11 & 17.89\\
w/o truncation & 0.14 & 1.29 & 2.06 & 0.14 & 6.28\\
\midrule
\rowcolor{SphereTint}
\textbf{SphereTTC} & 0.22 & 1.40 & 2.16 & 0.16 & 0.28\\
\bottomrule
\end{tabularx}
\end{table}

\subsection{Ablation Study}
\label{sec:component_analysis}
\vspace{-1mm}

Table~\ref{tab:component_analysis} compares SphereTTC with six variants on Z500 (upper-air circulation), T850 (lower-tropospheric temperature), and T2M (near-surface temperature), as defined in the Appendix. Each variant ablates one component:
(1)~\textbf{w/o gate} uses a fixed maximum blend;
(2)~\textbf{w/o taper} removes boundary tapering;
(3)~\textbf{w/o identity ridge} drops the identity penalty;
(4)~\textbf{w/o recency} uses uniform weights;
(5)~\textbf{w/o SH} fits grid points directly; and
(6)~\textbf{w/o truncation} keeps the full spectrum.
Full SphereTTC leads across all metrics: dropping the gate, SH, taper, or truncation spikes degradation (6.28--25.59\% vs.\ 0.28\%) and cuts gains, while omitting recency or identity ridge weakens T2M gains (and ACC for \textbf{w/o recency}).


\section{Conclusion and Future Work}
\label{sec:conclusion}
\vspace{-1mm}
This paper presents SphereTTC, the first gradient-free test-time calibration framework for frozen weather forecasters under delayed ground truth. Coupling tapered spherical harmonic corrections, a closed-form affine update, and chronological gating, it uses zero neural parameters, lowers errors on all 11 ERA5 backbones, and suppresses degraded outcomes. Limitations are detailed in the Appendix. Future work will explore joint calibration of coupled variables, vector winds, and probabilistic ensembles. Finally, while our cross-backbone benchmark adopts daily-mean targets for a unified evaluation contract, extending SphereTTC to 6-hourly instantaneous fields is a natural next step for calibrating faster, sub-daily atmospheric dynamics.

{
    \small
    \bibliographystyle{ieeenat_fullname}
    \bibliography{main}
}

\clearpage
\onecolumn
\appendix
\setcounter{table}{0}
\renewcommand{\thetable}{S\arabic{table}}
\renewcommand{\theHtable}{S\arabic{table}}
\setcounter{figure}{0}
\renewcommand{\thefigure}{S\arabic{figure}}
\renewcommand{\theHfigure}{S\arabic{figure}}
\setcounter{equation}{0}
\renewcommand{\theequation}{S\arabic{equation}}
\renewcommand{\theHequation}{S\arabic{equation}}
\begingroup
\setlength{\parindent}{1em}
\setlength{\parskip}{0pt}
\raggedbottom
\clubpenalty=10000
\widowpenalty=10000
\captionsetup{skip=3pt,hypcap=false}
\begin{center}
{\large\bfseries Enhancing Medium-range Global Weather Forecasting\\with Spherical Harmonic Test-Time Calibration}\par
\vspace{4pt}{\large Appendix}
\end{center}
\vspace{-4pt}
\noindent\textbf{Guide to the appendix.}
Appendices~\ref{app:data}--\ref{app:implementation} document data construction, complete scores, metrics, and calibration settings.
Appendix~\ref{app:analysis_protocol} defines verification availability; Appendix~\ref{app:references} specifies baselines and the matched-feedback comparison.
Appendices~\ref{app:spatial_spectral}--\ref{app:matched_spectral_control} examine spatial scales, gating, and the affine correction.
\par\vspace{4pt}
\section{Data and Forecast Construction}
\label{app:data}
\vspace{-1mm}
The evaluation contains 730 daily initializations in 2018--2019 at 1-, 3-, 5-, 7-, and 10-day leads.
The 49 channels comprise Z, Q, T, U, and V at 50, 100, 200, 300, 500, 700, 850, 925, and 1000\,hPa, plus T2M, U10, V10, and MSLP; all use a common $121\times240$ grid.
The complete tables report all 2,695 paired backbone--channel--lead score cells.

\noindent\textbf{Humidity conversion.}
For FourCastNetV2~\cite{bonev2023sfno} and FuXi~\cite{chen2023fuxi}, the archived adapters derive specific humidity from the predicted relative humidity and temperature at the same pressure level and forecast time.
Writing temperature in kelvin as $T$, relative humidity in percent as $r$, and pressure in hPa as $P$, the implemented conversion is
\begin{equation}
\begin{aligned}
T_c&=T-273.15,\qquad
e_s=611.2\exp\!\left(\frac{17.67T_c}{T_c+243.5}\right),\\
e&=\frac{\operatorname{clip}(r,0,100)}{100}\,e_s,\qquad
q=\frac{0.622e}{100P-0.378e}.
\end{aligned}
\label{eq:app_rh_to_q}
\end{equation}
Here $e_s$ and $e$ are in Pa and $q$ is in kg\,kg$^{-1}$.
The conversion uses the liquid-water saturation formula at every temperature. Each adapter converts the native-grid forecasts at 00, 06, 12, and 18 UTC, remaps the four fields linearly, and averages them into the target-day mean.
The adapters map the native \texttt{r} and \texttt{t} channels by their matching pressure suffix and retain the nine listed evaluation levels.

\begin{center}
\begin{minipage}{\textwidth}
\captionof{table}{Humidity handling in the archived forecaster adapters. Both adapters apply Eq.~\ref{eq:app_rh_to_q} at six-hourly resolution.}
\label{tab:app_humidity_adapters}
\small\setlength{\tabcolsep}{3pt}\renewcommand{\arraystretch}{1.12}
\begin{tabularx}{\textwidth}{@{}p{91pt}p{120pt}XX@{}}
\toprule
Forecaster & Native channels & WB2 initialization RH & Forecast post-processing\\
\midrule
FourCastNetV2 & RH (\%), same-level predicted $T$ (K), 13 pressure levels & Fraction passed directly to checkpoint normalization & Denormalize, convert RH to $q$, remap, average daily\\
FuXi & RH (\%), same-level predicted $T$ (K), 13 pressure levels & Multiply fraction by 100 and clip to $[0,100]$ & Convert RH to $q$, remap, average daily\\
\bottomrule
\end{tabularx}
\end{minipage}
\end{center}

\noindent\textbf{Initialization units.}
WeatherBench~2~\cite{rasp2024weatherbench2} supplies fractional RH, whereas the released inference interfaces expect percent.
Table~\ref{tab:app_humidity_adapters} specifies each initialization path separately from forecast post-processing.

\noindent\textbf{Sensitivity to excluding humidity.}
Calibration improves aggregate nRMSE for all 11 backbones across the 40 non-humidity channels.
Reaggregating the same archived scores, with forecasts, dates, and normalization fixed, gives mean and median reductions of 2.06\% and 0.97\%.
Table~\ref{tab:app_humidity_sensitivity} compares these results with the full 49-channel evaluation.

\begin{center}
\begin{minipage}{\textwidth}
\captionof{table}{Relative nRMSE reductions (\%) from archived cell scores with and without specific humidity. All five forecast leads are included.}
\label{tab:app_humidity_sensitivity}
\centering\small
\begin{tabular*}{0.80\textwidth}{@{\extracolsep{\fill}}lrr@{}}
\toprule
Backbone or summary & All 49 channels & 40 non-humidity channels\\
\midrule
FourCastNetV2 & 19.82 & 4.78\\
FuXi & 9.80 & 0.76\\
Across 11 backbones (mean) & 4.20 & 2.06\\
Across 11 backbones (median) & 1.34 & 0.97\\
\bottomrule
\end{tabular*}
\end{minipage}
\end{center}

\clearpage
\section{Complete Main Results and Cross-backbone Analyses}
\label{app:backbones}
\vspace{-1mm}
Table~\ref{tab:all_backbones} summarizes all 11 backbones; the five tables below give per-channel, per-lead scores, with SphereTTC in bold.

\begin{center}
\begin{minipage}{\textwidth}
\captionof{table}{Forecasting performance across all 11 backbones. Lower nRMSE and higher ACC are better; gain is the relative nRMSE reduction.}
\label{tab:all_backbones}
\begingroup\fontsize{8}{9}\selectfont\setlength{\tabcolsep}{5pt}
\begin{tabular*}{\textwidth}{@{\extracolsep{\fill}}lccccc@{}}
\toprule
& \multicolumn{3}{c}{nRMSE $\downarrow$} & \multicolumn{2}{c}{ACC $\uparrow$}\\
\cmidrule(lr){2-4}\cmidrule(lr){5-6}
Backbone & Raw & SphereTTC & Gain (\%) & Raw & SphereTTC\\
\midrule
ConvLSTM~\cite{shi2015convolutional} & 0.4698 & 0.4653 & 0.96 & 0.3600 & 0.3722 \\
Transformer~\cite{vaswani2017attention} & 0.4638 & 0.4589 & 1.05 & 0.3881 & 0.3988 \\
FNO~\cite{FNO} & 0.4427 & 0.4403 & 0.55 & 0.4424 & 0.4515 \\
ViT~\cite{dosovitskiy2020image} & 0.4477 & 0.4465 & 0.27 & 0.4321 & 0.4365 \\
CirT~\cite{Cirt} & 0.4605 & 0.4601 & 0.10 & 0.4156 & 0.4194 \\
ClimODE~\cite{verma2024climode} & 0.4798 & 0.4734 & 1.34 & 0.3679 & 0.3759 \\
FourCastNetV2~\cite{bonev2023sfno} & 0.4500 & 0.3608 & 19.82 & 0.6882 & 0.6972 \\
OneForecast~\cite{gao2025oneforecast} & 0.3120 & 0.2994 & 4.06 & 0.7740 & 0.7736 \\
FuXi~\cite{chen2023fuxi} & 0.3253 & 0.2935 & 9.80 & 0.7912 & 0.7965 \\
Pangu~\cite{bi2023accurate} & 0.2983 & 0.2873 & 3.70 & 0.7889 & 0.7883 \\
GraphCast~\cite{lam2023learning} & 0.2917 & 0.2784 & 4.57 & 0.7988 & 0.8004 \\
\bottomrule
\end{tabular*}
\endgroup
\end{minipage}
\end{center}
\vspace{-5pt}
\par\vspace{3pt}\noindent\begin{minipage}{\textwidth}
\centering
\includegraphics[width=\textwidth]{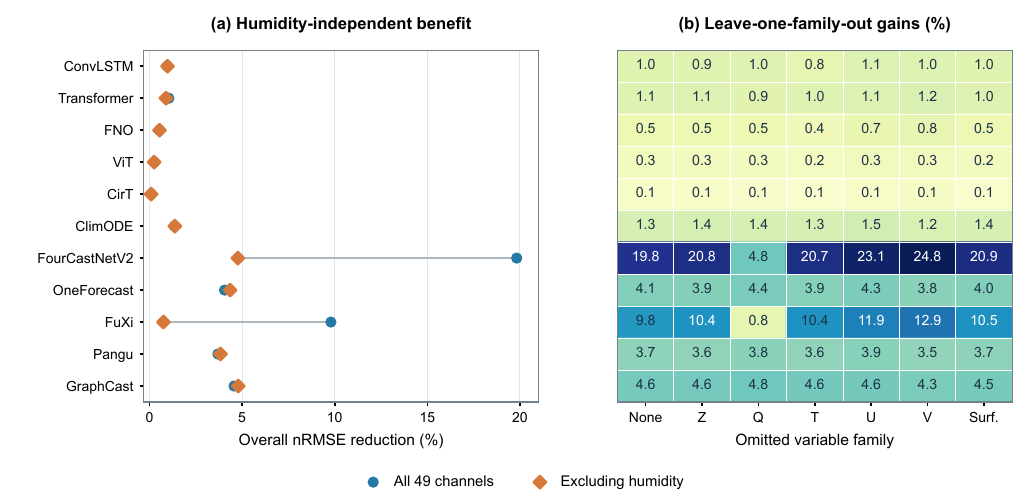}
\captionof{figure}{Sensitivity to the evaluated variable set. (a) Overall nRMSE reductions with and without humidity. (b) Gains after omitting each variable family; every backbone retains a positive aggregate gain.}
\label{fig:app_composition}
\end{minipage}\par\vspace{3pt}
\clearpage
\noindent\begin{minipage}{\textwidth}
\captionof{table}{Complete day-1 scores for 49 channels and 11 backbones. Each cell gives Raw / \textbf{SphereTTC} nRMSE (top) and ACC (bottom). Lower nRMSE and higher ACC are better.}
\label{tab:complete_day1}
\begingroup
\fontsize{6.5}{6.3}\selectfont
\setlength{\tabcolsep}{0pt}
\renewcommand{\arraystretch}{1}
\begin{tabular*}{\textwidth}{@{\extracolsep{\fill}}l*{11}{c}@{}}
\toprule
Channel & ConvLSTM~\cite{shi2015convolutional} & Transformer~\cite{vaswani2017attention} & FNO~\cite{FNO} & ViT~\cite{dosovitskiy2020image} & CirT~\cite{Cirt} & ClimODE~\cite{verma2024climode} & FourCastNetV2~\cite{bonev2023sfno} & OneForecast~\cite{gao2025oneforecast} & FuXi~\cite{chen2023fuxi} & Pangu~\cite{bi2023accurate} & GraphCast~\cite{lam2023learning} \\
\midrule
\rowcolor{black!3}
Z50 & \parbox[c][12.0pt][c]{41.5pt}{\centering 0.0902/\textbf{0.0770}\\[-0.1pt]0.888/\textbf{0.919}} & \parbox[c][12.0pt][c]{41.5pt}{\centering 0.0640/\textbf{0.0594}\\[-0.1pt]0.944/\textbf{0.952}} & \parbox[c][12.0pt][c]{41.5pt}{\centering 0.1032/\textbf{0.0930}\\[-0.1pt]0.847/\textbf{0.877}} & \parbox[c][12.0pt][c]{41.5pt}{\centering 0.0751/\textbf{0.0691}\\[-0.1pt]0.922/\textbf{0.934}} & \parbox[c][12.0pt][c]{41.5pt}{\centering 0.1041/\textbf{0.0971}\\[-0.1pt]0.843/\textbf{0.865}} & \parbox[c][12.0pt][c]{41.5pt}{\centering 0.0660/\textbf{0.0608}\\[-0.1pt]0.940/\textbf{0.949}} & \parbox[c][12.0pt][c]{41.5pt}{\centering 0.0384/\textbf{0.0250}\\[-0.1pt]0.980/\textbf{0.992}} & \parbox[c][12.0pt][c]{41.5pt}{\centering 0.0204/\textbf{0.0202}\\[-0.1pt]0.994/\textbf{0.995}} & \parbox[c][12.0pt][c]{41.5pt}{\centering 0.0201/\textbf{0.0188}\\[-0.1pt]0.995/\textbf{0.995}} & \parbox[c][12.0pt][c]{41.5pt}{\centering 0.0166/\textbf{0.0163}\\[-0.1pt]0.996/\textbf{0.996}} & \parbox[c][12.0pt][c]{41.5pt}{\centering 0.0243/\textbf{0.0242}\\[-0.1pt]0.992/\textbf{0.992}} \\
\bottomrule
\end{tabular*}
\endgroup
\end{minipage}
\clearpage
\noindent\begin{minipage}{\textwidth}
\captionof{table}{Complete day-3 scores for 49 channels and 11 backbones. Each cell gives Raw / \textbf{SphereTTC} nRMSE (top) and ACC (bottom). Lower nRMSE and higher ACC are better.}
\label{tab:complete_day3}
\begingroup
\fontsize{6.5}{6.3}\selectfont
\setlength{\tabcolsep}{0pt}
\renewcommand{\arraystretch}{1}
\begin{tabular*}{\textwidth}{@{\extracolsep{\fill}}l*{11}{c}@{}}
\toprule
Channel & ConvLSTM~\cite{shi2015convolutional} & Transformer~\cite{vaswani2017attention} & FNO~\cite{FNO} & ViT~\cite{dosovitskiy2020image} & CirT~\cite{Cirt} & ClimODE~\cite{verma2024climode} & FourCastNetV2~\cite{bonev2023sfno} & OneForecast~\cite{gao2025oneforecast} & FuXi~\cite{chen2023fuxi} & Pangu~\cite{bi2023accurate} & GraphCast~\cite{lam2023learning} \\
\midrule
\rowcolor{black!3}
Z50 & \parbox[c][12.0pt][c]{41.5pt}{\centering 0.1300/\textbf{0.1278}\\[-0.1pt]0.747/\textbf{0.756}} & \parbox[c][12.0pt][c]{41.5pt}{\centering 0.1151/\textbf{0.1148}\\[-0.1pt]0.805/\textbf{0.806}} & \parbox[c][12.0pt][c]{41.5pt}{\centering 0.1071/\textbf{0.1046}\\[-0.1pt]0.833/\textbf{0.842}} & \parbox[c][12.0pt][c]{41.5pt}{\centering 0.1054/\textbf{0.1046}\\[-0.1pt]0.839/\textbf{0.842}} & \parbox[c][12.0pt][c]{41.5pt}{\centering 0.1132/\textbf{0.1114}\\[-0.1pt]0.812/\textbf{0.819}} & \parbox[c][12.0pt][c]{41.5pt}{\centering 0.1264/\textbf{0.1257}\\[-0.1pt]0.765/\textbf{0.767}} & \parbox[c][12.0pt][c]{41.5pt}{\centering 0.0643/\textbf{0.0508}\\[-0.1pt]0.945/\textbf{0.965}} & \parbox[c][12.0pt][c]{41.5pt}{\centering 0.0409/\textbf{0.0407}\\[-0.1pt]0.978/\textbf{0.978}} & \parbox[c][12.0pt][c]{41.5pt}{\centering 0.0412/\textbf{0.0407}\\[-0.1pt]0.977/\textbf{0.978}} & \parbox[c][12.0pt][c]{41.5pt}{\centering 0.0324/\textbf{0.0323}\\[-0.1pt]0.986/\textbf{0.986}} & \parbox[c][12.0pt][c]{41.5pt}{\centering 0.0494/\textbf{0.0493}\\[-0.1pt]0.967/\textbf{0.967}} \\
\bottomrule
\end{tabular*}
\endgroup
\end{minipage}
\clearpage
\noindent\begin{minipage}{\textwidth}
\captionof{table}{Complete day-5 scores for 49 channels and 11 backbones. Each cell gives Raw / \textbf{SphereTTC} nRMSE (top) and ACC (bottom). Lower nRMSE and higher ACC are better.}
\label{tab:complete_day5}
\begingroup
\fontsize{6.5}{6.3}\selectfont
\setlength{\tabcolsep}{0pt}
\renewcommand{\arraystretch}{1}
\begin{tabular*}{\textwidth}{@{\extracolsep{\fill}}l*{11}{c}@{}}
\toprule
Channel & ConvLSTM~\cite{shi2015convolutional} & Transformer~\cite{vaswani2017attention} & FNO~\cite{FNO} & ViT~\cite{dosovitskiy2020image} & CirT~\cite{Cirt} & ClimODE~\cite{verma2024climode} & FourCastNetV2~\cite{bonev2023sfno} & OneForecast~\cite{gao2025oneforecast} & FuXi~\cite{chen2023fuxi} & Pangu~\cite{bi2023accurate} & GraphCast~\cite{lam2023learning} \\
\midrule
\rowcolor{black!3}
Z50 & \parbox[c][12.0pt][c]{41.5pt}{\centering 0.1505/\textbf{0.1490}\\[-0.1pt]0.644/\textbf{0.651}} & \parbox[c][12.0pt][c]{41.5pt}{\centering 0.1382/\textbf{0.1382}\\[-0.1pt]0.703/\textbf{0.703}} & \parbox[c][12.0pt][c]{41.5pt}{\centering 0.1253/\textbf{0.1244}\\[-0.1pt]0.763/\textbf{0.767}} & \parbox[c][12.0pt][c]{41.5pt}{\centering 0.1291/\textbf{0.1288}\\[-0.1pt]0.746/\textbf{0.748}} & \parbox[c][12.0pt][c]{41.5pt}{\centering 0.1290/\textbf{0.1284}\\[-0.1pt]0.749/\textbf{0.752}} & \parbox[c][12.0pt][c]{41.5pt}{\centering 0.1494/\textbf{0.1489}\\[-0.1pt]0.651/\textbf{0.653}} & \parbox[c][12.0pt][c]{41.5pt}{\centering 0.0864/\textbf{0.0758}\\[-0.1pt]0.900/\textbf{0.921}} & \parbox[c][12.0pt][c]{41.5pt}{\centering 0.0623/\textbf{0.0620}\\[-0.1pt]0.949/\textbf{0.949}} & \parbox[c][12.0pt][c]{41.5pt}{\centering 0.0606/\textbf{0.0605}\\[-0.1pt]0.950/\textbf{0.950}} & \parbox[c][12.0pt][c]{41.5pt}{\centering 0.0510/\textbf{0.0510}\\[-0.1pt]0.965/\textbf{0.965}} & \parbox[c][12.0pt][c]{41.5pt}{\centering 0.0698/\textbf{0.0697}\\[-0.1pt]0.934/\textbf{0.934}} \\
\bottomrule
\end{tabular*}
\endgroup
\end{minipage}
\clearpage
\noindent\begin{minipage}{\textwidth}
\captionof{table}{Complete day-7 scores for 49 channels and 11 backbones. Each cell gives Raw / \textbf{SphereTTC} nRMSE (top) and ACC (bottom). Lower nRMSE and higher ACC are better.}
\label{tab:complete_day7}
\begingroup
\fontsize{6.5}{6.3}\selectfont
\setlength{\tabcolsep}{0pt}
\renewcommand{\arraystretch}{1}
\begin{tabular*}{\textwidth}{@{\extracolsep{\fill}}l*{11}{c}@{}}
\toprule
Channel & ConvLSTM~\cite{shi2015convolutional} & Transformer~\cite{vaswani2017attention} & FNO~\cite{FNO} & ViT~\cite{dosovitskiy2020image} & CirT~\cite{Cirt} & ClimODE~\cite{verma2024climode} & FourCastNetV2~\cite{bonev2023sfno} & OneForecast~\cite{gao2025oneforecast} & FuXi~\cite{chen2023fuxi} & Pangu~\cite{bi2023accurate} & GraphCast~\cite{lam2023learning} \\
\midrule
\rowcolor{black!3}
Z50 & \parbox[c][12.0pt][c]{41.5pt}{\centering 0.1616/\textbf{0.1598}\\[-0.1pt]0.578/\textbf{0.588}} & \parbox[c][12.0pt][c]{41.5pt}{\centering 0.1503/\textbf{0.1502}\\[-0.1pt]0.636/\textbf{0.636}} & \parbox[c][12.0pt][c]{41.5pt}{\centering 0.1415/\textbf{0.1408}\\[-0.1pt]0.686/\textbf{0.689}} & \parbox[c][12.0pt][c]{41.5pt}{\centering 0.1441/\textbf{0.1439}\\[-0.1pt]0.671/\textbf{0.672}} & \parbox[c][12.0pt][c]{41.5pt}{\centering 0.1421/\textbf{0.1426}\\[-0.1pt]0.685/\textbf{0.685}} & \parbox[c][12.0pt][c]{41.5pt}{\centering 0.1635/\textbf{0.1626}\\[-0.1pt]0.575/\textbf{0.579}} & \parbox[c][12.0pt][c]{41.5pt}{\centering 0.1118/\textbf{0.1024}\\[-0.1pt]0.833/\textbf{0.853}} & \parbox[c][12.0pt][c]{41.5pt}{\centering 0.0880/\textbf{0.0871}\\[-0.1pt]0.898/\textbf{0.899}} & \parbox[c][12.0pt][c]{41.5pt}{\centering 0.0817/\textbf{0.0818}\\[-0.1pt]0.908/\textbf{0.908}} & \parbox[c][12.0pt][c]{41.5pt}{\centering 0.0754/\textbf{0.0754}\\[-0.1pt]0.924/\textbf{0.924}} & \parbox[c][12.0pt][c]{41.5pt}{\centering 0.0913/\textbf{0.0914}\\[-0.1pt]0.886/\textbf{0.886}} \\
\bottomrule
\end{tabular*}
\endgroup
\end{minipage}
\clearpage
\noindent\begin{minipage}{\textwidth}
\captionof{table}{Complete day-10 scores for 49 channels and 11 backbones. Each cell gives Raw / \textbf{SphereTTC} nRMSE (top) and ACC (bottom). Lower nRMSE and higher ACC are better.}
\label{tab:complete_day10}
\begingroup
\fontsize{6.5}{6.3}\selectfont
\setlength{\tabcolsep}{0pt}
\renewcommand{\arraystretch}{1}
\begin{tabular*}{\textwidth}{@{\extracolsep{\fill}}l*{11}{c}@{}}
\toprule
Channel & ConvLSTM~\cite{shi2015convolutional} & Transformer~\cite{vaswani2017attention} & FNO~\cite{FNO} & ViT~\cite{dosovitskiy2020image} & CirT~\cite{Cirt} & ClimODE~\cite{verma2024climode} & FourCastNetV2~\cite{bonev2023sfno} & OneForecast~\cite{gao2025oneforecast} & FuXi~\cite{chen2023fuxi} & Pangu~\cite{bi2023accurate} & GraphCast~\cite{lam2023learning} \\
\midrule
\rowcolor{black!3}
Z50 & \parbox[c][12.0pt][c]{41.5pt}{\centering 0.1745/\textbf{0.1726}\\[-0.1pt]0.497/\textbf{0.509}} & \parbox[c][12.0pt][c]{41.5pt}{\centering 0.1625/\textbf{0.1627}\\[-0.1pt]0.559/\textbf{0.559}} & \parbox[c][12.0pt][c]{41.5pt}{\centering 0.1634/\textbf{0.1616}\\[-0.1pt]0.565/\textbf{0.574}} & \parbox[c][12.0pt][c]{41.5pt}{\centering 0.1589/\textbf{0.1592}\\[-0.1pt]0.583/\textbf{0.582}} & \parbox[c][12.0pt][c]{41.5pt}{\centering 0.1577/\textbf{0.1600}\\[-0.1pt]0.593/\textbf{0.591}} & \parbox[c][12.0pt][c]{41.5pt}{\centering 0.1790/\textbf{0.1780}\\[-0.1pt]0.476/\textbf{0.482}} & \parbox[c][12.0pt][c]{41.5pt}{\centering 0.1483/\textbf{0.1371}\\[-0.1pt]0.708/\textbf{0.728}} & \parbox[c][12.0pt][c]{41.5pt}{\centering 0.1295/\textbf{0.1258}\\[-0.1pt]0.782/\textbf{0.785}} & \parbox[c][12.0pt][c]{41.5pt}{\centering 0.1141/\textbf{0.1138}\\[-0.1pt]0.815/\textbf{0.816}} & \parbox[c][12.0pt][c]{41.5pt}{\centering 0.1182/\textbf{0.1159}\\[-0.1pt]0.814/\textbf{0.816}} & \parbox[c][12.0pt][c]{41.5pt}{\centering 0.1299/\textbf{0.1279}\\[-0.1pt]0.770/\textbf{0.772}} \\
\bottomrule
\end{tabular*}
\endgroup
\end{minipage}

\clearpage
\subsection{Expanded Lead-time Curves}
\vspace{-1mm}
Figures~\ref{fig:app_local_leads} and~\ref{fig:app_pretrained_leads} resolve five physical variables for all 11 backbones.
Raw and calibrated forecasts share the same evaluation cases within each backbone; local and released systems are separated to keep their error ranges readable.
Showing RMSE and ACC together distinguishes changes in error magnitude from changes in anomaly agreement.
\par\vspace{3pt}\noindent\begin{minipage}{\textwidth}
\centering
\includegraphics[width=\textwidth]{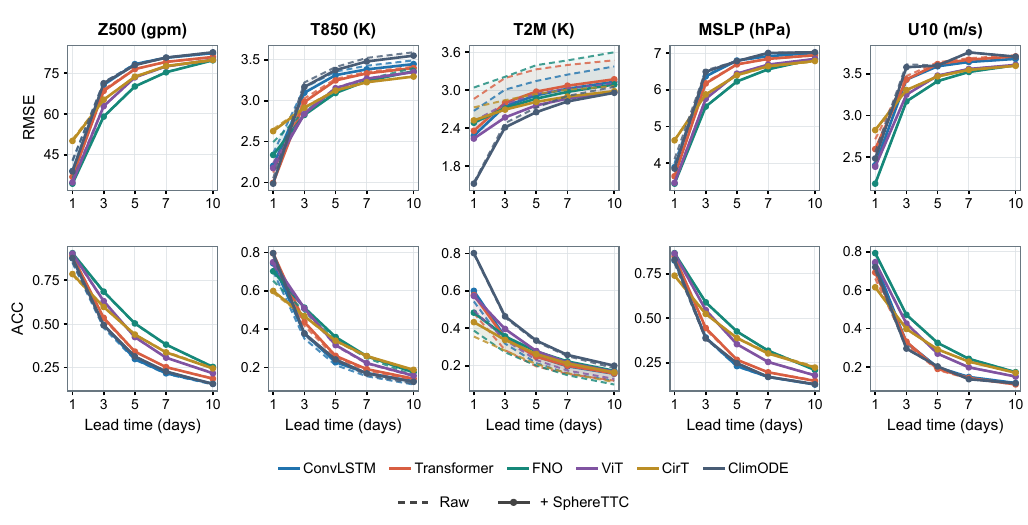}
\captionof{figure}{Lead-time RMSE and ACC for six locally trained backbones. Dashed: Raw; solid with markers: SphereTTC.}
\label{fig:app_local_leads}
\end{minipage}\par\vspace{3pt}
\vspace{-3pt}
\par\vspace{3pt}\noindent\begin{minipage}{\textwidth}
\centering
\includegraphics[width=\textwidth]{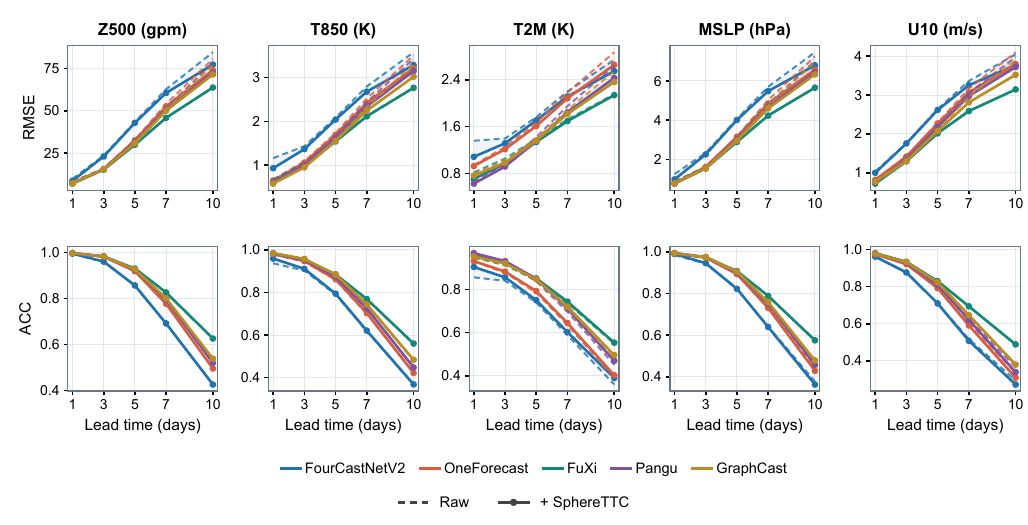}
\captionof{figure}{Lead-time RMSE and ACC for the five released forecasting systems, using the same line conventions as Figure~\ref{fig:app_local_leads}.}
\label{fig:app_pretrained_leads}
\end{minipage}\par\vspace{3pt}

\clearpage
\subsection{Variable and Pressure-level Coverage}
\vspace{-1mm}
Figure~\ref{fig:app_channel_gains} retains all 49 channels and averages the five lead scores equally before computing changes.
At individual leads, RMSE improves in 2,407 of 2,695 cells and ACC in 2,135.
The signed maps show how these gains vary across variable families, pressure levels, and forecasting architectures.
\par\vspace{3pt}\noindent\begin{minipage}{\textwidth}
\centering
\includegraphics[width=\textwidth]{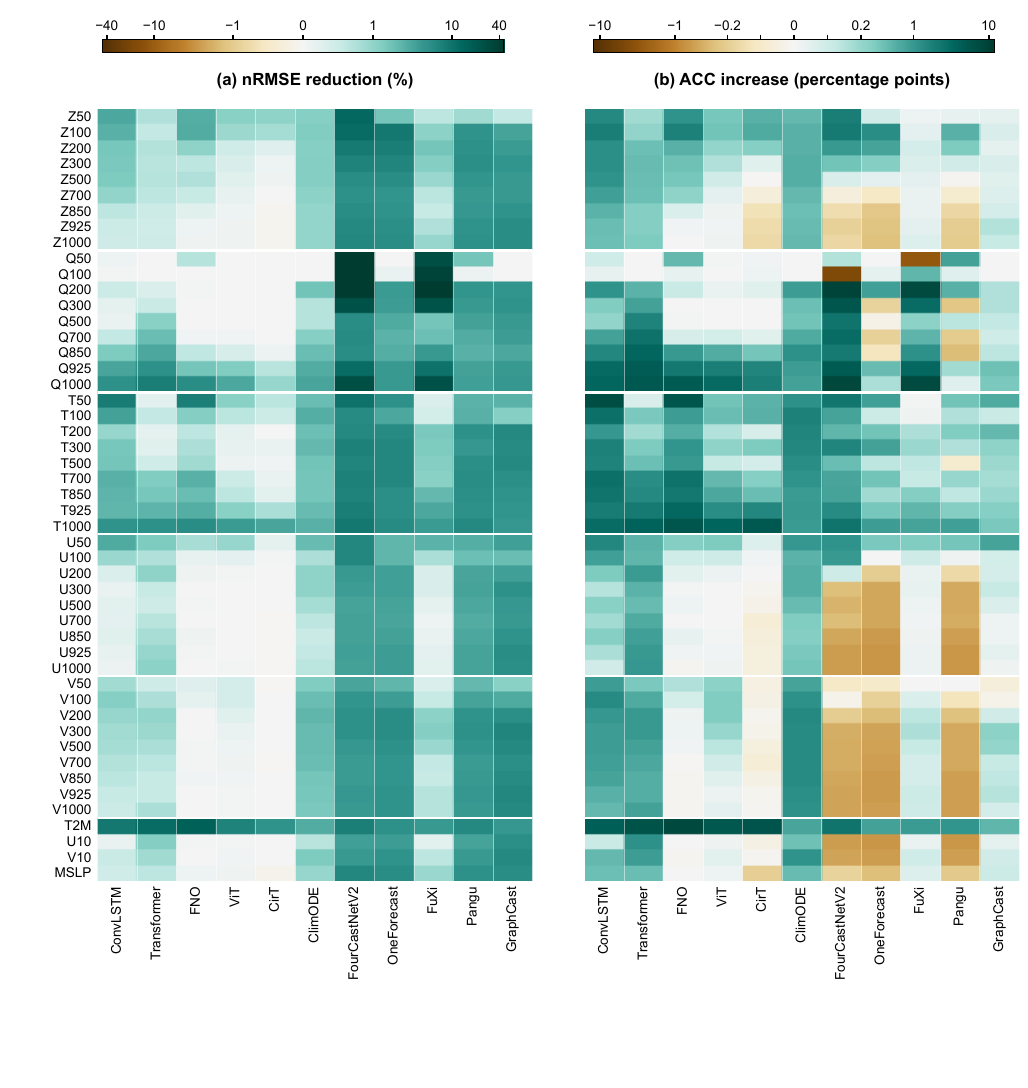}
\captionof{figure}{Channel-wise calibration effects across all 11 backbones. Teal denotes improvement and brown deterioration; the symmetric-log color scales preserve both small effects and large humidity corrections. ACC changes are absolute percentage-point differences.}
\label{fig:app_channel_gains}
\end{minipage}\par\vspace{3pt}

\clearpage
\subsection{Transfer to a Subsequent Year}
\vspace{-1mm}
The GraphCast~\cite{lam2023learning} assessment uses 366 initializations in 2020, with one configuration selected from 56 candidates on two 2017 development segments.
The configuration remains fixed throughout all 245 channel--lead comparisons, assessing transfer to a later year without reselecting calibration hyperparameters.
\par\vspace{3pt}\noindent\begin{minipage}{\textwidth}
\centering
\includegraphics[width=\textwidth]{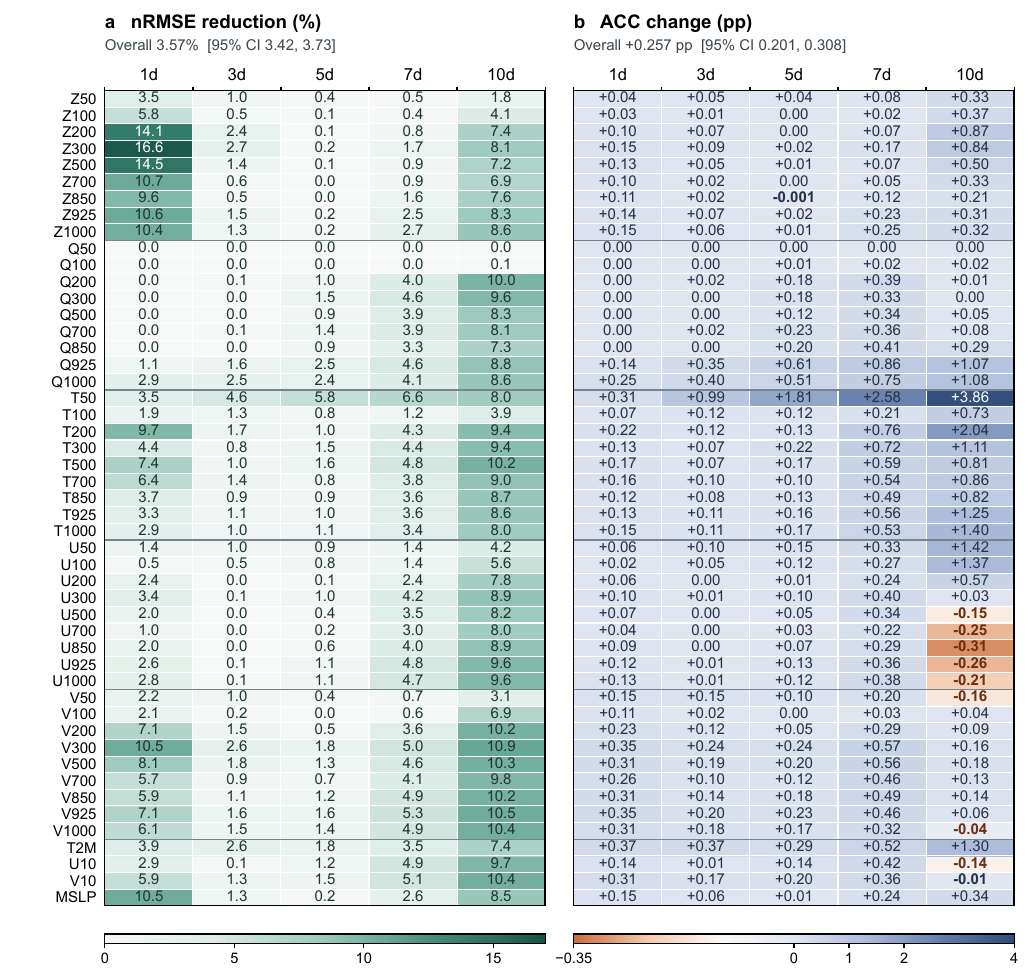}
\captionof{figure}{GraphCast~\cite{lam2023learning} calibration effects in 2020 across all 49 channels and five leads. Calibration settings are frozen before scoring; verification is admitted at its target-date timestamp, $t_{\mathrm{valid},s}(\tau)\leq t$.}
\label{fig:app_frozen2020}
\end{minipage}\par\vspace{3pt}
\noindent The 14-day block-bootstrap 95\% intervals are [3.424, 3.729]\% for mean cell-relative RMSE gain and [0.201, 0.308] percentage points for ACC change.

\section{Metrics and Aggregation}
\label{app:metrics}
\vspace{-1mm}
For normalized cosine-latitude weights $w_i$, the pooled error and anomaly correlation of one channel and lead are
\begin{equation}
R=\left[\frac{1}{N}\sum_{t,i}w_i(\widehat y_{t,i}-y_{t,i})^2\right]^{1/2},\qquad
A=\frac{\sum_{t,i}w_i a_{t,i}b_{t,i}}
{\sqrt{\sum_{t,i}w_i a_{t,i}^2\sum_{t,i}w_i b_{t,i}^2}}.
\end{equation}
Here $N$ is the number of initializations, and $a$ and $b$ are the predicted and observed anomalies.
The macro nRMSE averages $R_{v,\tau}/\sigma_v$ equally across channels and leads, using channel standard deviations from 1979--2016.
These scoring scales are separate from each released forecaster's native input normalization.
Main-evaluation anomalies use the 1979--2016 calendar-day climatology; diagnostic anomalies use the WeatherBench~2 1959--2015 climatology.
Relative score changes are calculated after macro aggregation:
\begin{equation}
\Delta E=100(1-E_{\rm cal}/E_{\rm raw}),\qquad
\Delta A_{\rm rel}=100(A_{\rm cal}/A_{\rm raw}-1).
\label{eq:relative_score_change}
\end{equation}
In the main paper, Table~\ref{tab:main_results} shows magnitudes with direction arrows, whereas Tables~\ref{tab:calibrator_comparison} and~\ref{tab:component_analysis} use signed improvements.
The diagnostic plots report $100(A_{\rm cal}-A_{\rm raw})$ in percentage points to avoid dividing by nonpositive individual correlations.
The 2020 summary reports the mean of cell-relative RMSE gains, which differs from the main macro ratio.

\section{Implementation and Calibration Settings}
\label{app:implementation}
\vspace{-1mm}
Six local backbones are trained on 1979--2016 initializations using four NVIDIA H100 80G GPUs, with 2017 for validation; five systems use released checkpoints.
All forecasting weights remain fixed during calibration; architecture and checkpoint sources are detailed in Appendix~\ref{app:forecaster_baselines}.
Local models output daily fields; released systems construct daily means from the 00, 06, 12, and 18 UTC forecast states on each target day and express their outputs on the common grid.

\begin{center}
\captionof{table}{SphereTTC settings. $K$: memory size; $h$: recency half-life; $L$: exclusive harmonic bandwidth; $\gamma$: maximum strength.}
\label{tab:settings}
\small\setlength{\tabcolsep}{5pt}
\begin{tabular*}{\textwidth}{@{\extracolsep{\fill}}llrrrr@{}}
\toprule Backbone & Forecasting implementation & $K$&$h$&$L$&$\gamma$\\\midrule
ConvLSTM~\cite{shi2015convolutional}, Transformer~\cite{vaswani2017attention}&Locally trained architectures&64&32&40&1.0\\
FNO~\cite{FNO}, ViT~\cite{dosovitskiy2020image}&Locally trained architectures&64&32&40&1.0\\
CirT~\cite{Cirt}&Geometry-inspired core; common output head&32&16&32&1.0\\
ClimODE~\cite{verma2024climode}&Local neural ODE with advection and Euler updates&64&32&40&1.0\\
FourCastNetV2~\cite{bonev2023sfno}, OneForecast~\cite{gao2025oneforecast}&Released pretrained checkpoints&64&32&24&1.0\\
FuXi~\cite{chen2023fuxi}, Pangu~\cite{bi2023accurate}&Released pretrained checkpoints&64&32&24&1.0\\
GraphCast~\cite{lam2023learning}&Released smaller $1^\circ$ checkpoint&64&32&64&0.8\\
\bottomrule
\end{tabular*}
\end{center}
\noindent All backbones share ridge multiplier $\lambda_0=0.05$, reliability penalty $z=0.5$, residual cap $\kappa=3$, scale bound $\delta=0.5$, and taper exponent $p=1$.
The minimum eligible history is 24 pairs; chronological validation reserves the most recent quarter, with at least eight pairs.
Ten of the 11 backbones share $(K,h)=(64,32)$, with only $(L,\gamma)$ selected on 2017 verification before testing.
The main and recorded 2020 GraphCast evaluations use GraphCast\_small, whose documented training period is 1979--2015.
Appendix~\ref{app:forecaster_baselines} details architectures and released-weight provenance; Appendix~\ref{app:analysis_protocol} gives verification schedules.

\noindent Neural parameter counts include all neural weights; SphereTTC has zero neural parameters but retains fitted harmonic coefficients and stored history.

\section{Verification Availability and Analysis Cohorts}
\label{app:analysis_protocol}
\vspace{-1mm}
Let $t_{\mathrm{valid},s}(\tau)$ denote the start of the target day for initialization $s$ and lead $\tau$.
Completed-day verification becomes eligible after that day and any additional reporting delay $d$ have elapsed:
\begin{equation}
 r_s(\tau)=t_{\mathrm{valid},s}(\tau)+24\,\mathrm{h}+d,\qquad r_s(\tau)\leq t,
 \label{eq:verification_availability}
\end{equation}

Here $t$ is the current forecast issuance time; $d=0$ admits verification immediately after the target day ends.
Released systems issue forecasts at 00\,UTC on the initialization day; locally trained models issue them after their daily-mean input day is complete.
For completed-day replays, Eq.~\ref{eq:verification_availability} governs fitting, gate validation, warmup, online comparator updates, and development-period hyperparameter selection.
The archived main and 2020 evaluations use target-date admission, comparing target-day timestamps with the recorded initialization timestamp (Table~\ref{tab:app_cohorts}).

\begin{center}
\begin{minipage}{\textwidth}
\captionof{table}{Evaluation cohorts and verification availability. $d$ is an additional verification delay.}
\label{tab:app_cohorts}
\small\setlength{\tabcolsep}{4pt}
\begin{tabularx}{\textwidth}{@{}lclX@{}}
\toprule Cohort & Dates & Coverage & Verification availability\\\midrule
Main evaluation &730&11 models, 49 channels, 5 leads&Target-date admission across the archived 730-origin evaluation\\
GraphCast~\cite{lam2023learning} 2020 &366&49 channels, 5 leads&Target-date admission ($t_{\mathrm{valid},s}(\tau)\leq t$); settings frozen before testing\\
Matched diagnostic &287&2 models, 5 channels, 4 leads&Completed daily mean; 13 March--24 December 2018\\
Matched comparison/control &238&1 May--24 December 2018&Completed daily mean; original comparison validation ends 22 April, followed by eight unscored days\\
Spatial diagnostics &299&Analysis-specific coverage&Separate 1 March--24 December 2018 diagnostic cohort\\
\bottomrule\end{tabularx}
\end{minipage}
\end{center}
The component and delay replays construct WeatherBench~2 daily targets from 00, 06, 12, and 18 UTC.
Within each matched comparison, methods share forecast fields, targets, and evaluation dates; their admission schedules are specified in Appendix~\ref{app:references}.
The main evaluation uses the daily-field construction in Appendix~\ref{app:implementation}.

\subsection{Matched Component Replay}
\label{app:component_replay}
\vspace{-1mm}
The completed-day replay examines the balance between aggregate error reduction and individual forecast deterioration over 287 dates, five variables, and four leads.
Each backbone contributes 5,740 initialization--variable--lead outcomes.
Gating reduces degraded outcomes from 1,469 to 16 for GraphCast~\cite{lam2023learning} and from 1,026 to 59 for Pangu~\cite{bi2023accurate}, with smaller mean nRMSE gains (Figure~\ref{fig:reliability_cost}(a)).
Panel (b) isolates feedback age using 282 common GraphCast dates and 64 eligible pairs at every delay.
Positive nRMSE gains persist throughout this delay range.

\par\vspace{3pt}\noindent\begin{minipage}{\textwidth}\centering
\includegraphics[width=\textwidth]{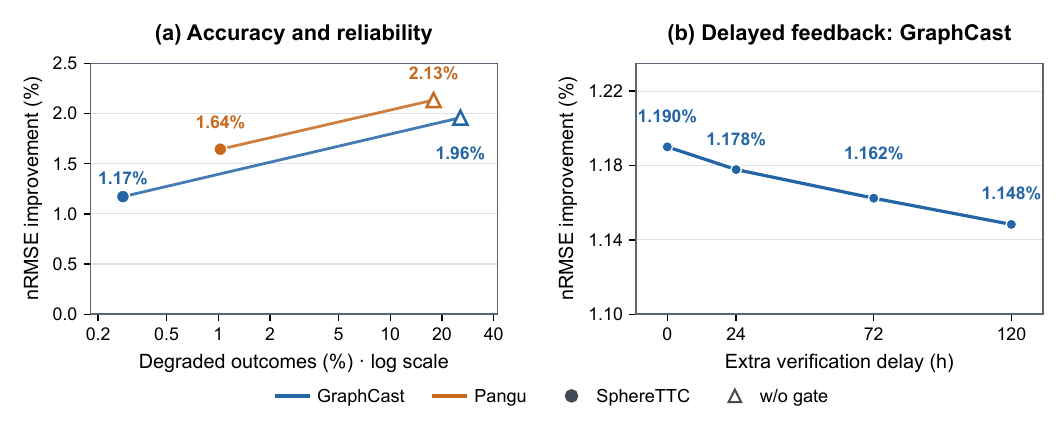}
\captionof{figure}{Calibration trade-offs and sensitivity to delayed feedback.}
\label{fig:reliability_cost}
\end{minipage}\par\vspace{2pt}

\noindent\begin{minipage}[t]{0.46\textwidth}
\noindent\textbf{Temporal variation.}\label{app:temporal_variation}
The main-text 84-day window maximizes the mean global-mean-series gain among 204 eligible windows and is illustrative.
Each point is a separately initialized forecast rather than a long rollout; no smoothing is applied.

Figure~\ref{fig:temporal_full_cohort} retains the complete matched sequence at five days, with mean-series RMSE reductions of 16.7\% (T2M) and 8.5\% (T850).
These series errors differ from spatial forecast RMSE.
\end{minipage}\hfill
\begin{minipage}[t]{0.51\textwidth}\centering\vspace{0pt}
\includegraphics[width=\linewidth]{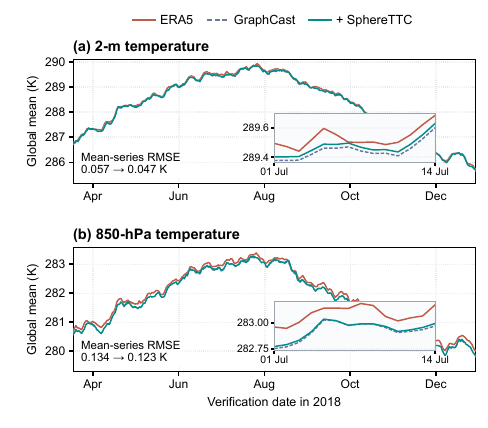}
\captionof{figure}{Full-cohort GraphCast~\cite{lam2023learning} temperature variation at five days. Insets retain the same July interval.}
\label{fig:temporal_full_cohort}
\end{minipage}

\section{Forecasting and Calibration Baselines}
\label{app:references}
\vspace{-1mm}
\subsection{Forecasters and Weight Sources}
\label{app:forecaster_baselines}
\vspace{-1mm}
The six local architectures in Table~\ref{tab:forecaster_baselines} are trained from random initialization on 1979--2016 ERA5 data and validated on 2017.
The five released systems use the checkpoints and native normalization files listed in Table~\ref{tab:released_forecaster_weights}.
Forecasting weights remain fixed throughout calibration.

\begin{center}
\begin{minipage}{\textwidth}
\captionof{table}{Locally trained forecasting architectures.}
\label{tab:forecaster_baselines}
\begingroup\fontsize{8.5}{10}\selectfont
\setlength{\tabcolsep}{4pt}\renewcommand{\arraystretch}{1.12}
\begin{tabularx}{\textwidth}{@{}p{100pt}X@{}}
\toprule
Forecaster & Architecture and implementation\\
\midrule
ConvLSTM~\cite{shi2015convolutional} & Convolutional encoder, recurrent convolutional LSTM state, and leadwise field decoder.\\
Transformer~\cite{vaswani2017attention} & Grid tokenization and global self-attention, followed by interpolation of the output fields.\\
FNO~\cite{FNO} & Truncated Fourier convolutions and pointwise layers, with a joint output head for the evaluated leads.\\
ViT~\cite{dosovitskiy2020image} & Patch tokenization and transformer encoding, followed by a learned field reconstruction head.\\
CirT~\cite{Cirt} & Official geometry-inspired core with the final head adapted to the common variables and leads; the adapted network is trained locally.\\
ClimODE~\cite{verma2024climode} & Local neural-ODE implementation with an advective component and Euler updates on the common grid.\\
\bottomrule\end{tabularx}\endgroup
\end{minipage}
\end{center}

\begin{center}
\begin{minipage}{\textwidth}
\captionof{table}{Forecast assets, reported training and selection periods, and their relation to the evaluation cohorts.}
\label{tab:released_forecaster_weights}
\begingroup\fontsize{8.5}{10}\selectfont
\setlength{\tabcolsep}{4pt}\renewcommand{\arraystretch}{1.12}
\begin{tabularx}{\textwidth}{@{}p{90pt}p{190pt}X@{}}
\toprule
Forecaster & Asset named in setup & Documented training and selection\\
\midrule
Six local models & Locally configured checkpoints, initialized from scratch & Training 1979--2016; validation 2017; 2018--2019 origins are outside the configured development period.\\
FourCastNetV2~\cite{bonev2023sfno} & \href{https://github.com/ecmwf-lab/ai-models-fourcastnetv2}{ECMWF inference package}: \texttt{weights.tar}, \texttt{global\_means.npy}, and \texttt{global\_stds.npy}. & SFNO reports training on 1979--2015 and selection on 2016--2017.\\
OneForecast~\cite{gao2025oneforecast} & \href{https://huggingface.co/YuanGao-YG/OneForecast/tree/main}{Author release}: \texttt{best\_ckpt.tar}, experiment \texttt{20241008-171138}, with its configuration and normalization files. & Published training spans 1959--2017 and selection uses 2018--2019. Those selection years overlap this cohort.\\
FuXi~\cite{chen2023fuxi} & \href{https://doi.org/10.5281/zenodo.10401602}{Author release}: \texttt{short.onnx}, \texttt{medium.onnx}, \texttt{long.onnx}, and external tensor files. & The paper reports 1979--2015 pretraining, 2016--2017 validation, and 2012--2017 forecasts for cascade fine-tuning.\\
Pangu~\cite{bi2023accurate} & \href{https://github.com/198808xc/Pangu-Weather}{Author release}: \texttt{pangu\_weather\_24.onnx} and \texttt{pangu\_weather\_6.onnx}. & Published training spans 1979--2017, with 2019 used for validation. The reported validation year overlaps this cohort.\\
GraphCast~\cite{lam2023learning} & \href{https://github.com/google-deepmind/weathernext/blob/master/docs/weathernext1_graph/README.md}{GraphCast\_small}, $1^{\circ}$, 13 pressure levels; local alias \texttt{GraphCast\_small\_1p0deg.npz}. & The released small model is documented as trained on ERA5 1979--2015; the recorded 2020 run includes its checkpoint hash. Main and 2020 targets fall after the reported training years.\\
GraphCast 2020 & Locally generated forecasts from GraphCast\_small; separate 2019 warm-up cache. & Same documented 1979--2015 training period, with a recorded checkpoint hash.\\
WB2 GraphCast 2018 & Precomputed \href{https://weatherbench2.readthedocs.io/en/latest/data-guide.html}{WeatherBench~2} forecast archive. & WB2 reports training through 2017 for the 2018 archive.\\
WB2 Pangu 2018 & Precomputed WeatherBench~2 2018 forecasts from the public Pangu implementation. & Forecast origins and targets in this replay fall in 2018, outside Pangu's reported 1979--2017 training and 2019 validation years.\\
\bottomrule\end{tabularx}\endgroup
\end{minipage}
\end{center}

\noindent\textbf{Checkpoint handling.}
Table~\ref{tab:released_forecaster_weights} distinguishes the configured local-model split from the published training and selection periods of released models.
The recorded 2020 GraphCast\_small checkpoint has SHA-256 prefix \texttt{e9438d8ad31ca6e1d3397a33}.
Its adapter loads \texttt{mean\_by\_level.nc}, \texttt{stddev\_by\_level.nc}, and \texttt{diffs\_stddev\_by\_level.nc}.

The main cohort contains 730 origins in 2018--2019, with target days from 2 January 2018 through 10 January 2020.
These evaluation years overlap OneForecast's published selection period and Pangu's 2019 validation period (Table~\ref{tab:released_forecaster_weights}).
OneForecast uses 2020 as its held-out year; the matched 238-origin Pangu replay scores 2018 targets, outside its reported training and validation periods.
The separate GraphCast 2020 assessment contains 366 origins, with target days from 2 January 2020 through 10 January 2021.

\noindent\textbf{Analytical forecast archives.}
The diagnostic comparisons on GraphCast~\cite{lam2023learning} and Pangu~\cite{bi2023accurate} replay published WeatherBench~2~\cite{rasp2024weatherbench2} forecast arrays.
Calibration and scoring use the saved predictions directly.
The calibrators share the same archive versions and initializations. The matched 238-origin comparison covers 1 May--24 December 2018 with target days from 2 May to 31 December 2018.

\subsection{Calibration Baselines and Initialization}
\label{app:calibration_baselines}
\vspace{-1mm}
The calibration baselines comprise four training-time networks, four gradient-based test-time adapters, and two statistical methods.
Tables~\ref{tab:app_calibrators} and~\ref{tab:calibrator_weight_sources} describe their mechanisms, implementations, and initialization.
The common interface operates on five forecast channels and four leads on the $121\times240$ grid.
It evaluates output calibration modules; input calibration and backbone execution are excluded.

\noindent\textbf{Training-time calibration.}
The regional, station, or coupled-system interfaces are adapted to this common global workload.
For the accuracy comparison in Table~\ref{tab:calibrator_comparison} of the main paper, all four networks are trained offline on the pre-embargo 2018 verified archive ending 22 April 2018 (with chronological validation for model selection prior to the eight-day embargo) and kept frozen throughout the 238-day test period.
SphereTTC updates its coefficients in closed form as eligible verification arrives.
For the cost measurements in Table~\ref{tab:calibration_cost} of the main paper, the four cores retain initialized weights and no pretrained calibration checkpoint is loaded; offline fitting is excluded.
The source revisions below identify implementations, rather than trained weight files.

\begin{center}
\begin{minipage}{\textwidth}
\captionof{table}{Training-time calibration architectures and implementation sources.}
\label{tab:app_calibrators}
\begingroup\fontsize{8.5}{10}\selectfont
\setlength{\tabcolsep}{4pt}\renewcommand{\arraystretch}{1.12}
\begin{tabularx}{\textwidth}{@{}p{90pt}X p{165pt}@{}}
\toprule
Calibrator & Implemented output interface & Implementation source and cost-test weights\\
\midrule
DRN~\cite{rasp2018neuralpostprocessing} & Gaussian location/scale heads; the predictive mean supplies the deterministic output. The common-grid adaptation uses location embeddings, geographic features, and forecast lead as inputs~\cite{demaeyer2023euppbench}. & \href{https://github.com/EUPP-benchmark/ESSD-DRN}{EUPP benchmark}; adapted PyTorch interface. Reference-file SHA prefix \texttt{091080e}; initialized weights.\\
BC-UNet~\cite{cui2025bcunet} & Released SmaAt-UNet~\cite{trebing2021smaatunet} encoder--decoder with attention, depthwise-separable convolutions, and skip connections; forecasts map directly to corrected fields. & \href{https://github.com/NOAA-EMC/ML4BC}{NOAA-EMC/ML4BC}, revision \texttt{41dfede}; initialized five-channel core.\\
PnP-Corrector~\cite{pnpcorrector} & External correction network around a frozen forecaster; the released core retains its native layer widths and block counts. & \href{https://github.com/Alexander-wu/PnP-Corrector}{Author implementation}, revision \texttt{70e63ba}; initialized correction core.\\
ASTAFN~\cite{xu2026astafn} & Spatiotemporal alignment and graph fusion of forecasts and observations. The grid interface uses all 29,040 nodes, four completed observation days, five target heads, and four-neighbor geographic support periodic in longitude. & \href{https://github.com/xubihe-bjtu/ASTAFN}{Author implementation}, revision \texttt{987adf1}; initialized target cores.\\
\bottomrule\end{tabularx}\endgroup
\end{minipage}
\end{center}

\noindent\textbf{Gradient-based test-time calibration.}
All four adapters use their released identity-preserving initialization, without a pretrained calibration checkpoint.
Their weights are learned chronologically with one Adam update for each newly completed four-lead verification window.
After learning-rate selection, model and optimizer states continue through the embargo and test periods without resetting.

\begin{center}
\begin{minipage}{\textwidth}
\captionof{table}{Online calibration mechanisms, source revisions, and initialization.}
\label{tab:calibrator_weight_sources}
\begingroup\fontsize{8.5}{10}\selectfont
\setlength{\tabcolsep}{4pt}\renewcommand{\arraystretch}{1.12}
\begin{tabularx}{\textwidth}{@{}p{85pt}X p{177pt}@{}}
\toprule
Calibrator & Calibration mechanism & Source revision and initial weights\\
\midrule
TAFAS~\cite{kim2025tafas} & Residual gated calibration with shared lead-mixing weights and coordinate-specific biases and gates. & \href{https://github.com/kimanki/TAFAS}{Author implementation}: \texttt{139bf98}. Zero temporal weights/biases; gate 0.01.\\
ST-TTC~\cite{chen2025stttc} & Amplitude and phase adjustments in temporal frequency; the four leads yield three real-FFT bins, each assigned a calibration group. & \href{https://github.com/Onedean/ST-TTC}{Author implementation}: \texttt{ddba3fe}. Zero amplitude/phase adjustments.\\
COSA~\cite{im2026cosa} & Shared gated linear output adapter, conditioned on one normalized target mean from each of the five latest fully verified forecast windows. Issuance-time context is reused at the later update. & \href{https://github.com/bigbases/COSA_ICLR2026}{Author implementation}: \texttt{527c0fe}. Zero gate and a small Xavier-initialized linear head.\\
PETSA~\cite{medeiros2025petsa} & Coordinate-wise low-rank output module, with rank 16 and the complete-label loss. & \href{https://github.com/BorealisAI/PETSA}{Author implementation}: \texttt{87853d8}. Zero low-rank output factor/bias; gate 0.01.\\
\bottomrule\end{tabularx}\endgroup
\end{minipage}
\end{center}

\noindent\textbf{Statistical post-processing.}
Running-mean bias correction adds the unweighted mean residual from the latest 64 eligible pairs, separately for each lead and grid cell.
Grid-point MOS~\cite{glahn1972mos} fits an affine map on the same memory, with an unpenalized intercept and a ridge penalty on the slope.
Both methods estimate their coefficients directly from verified history and use no neural-network checkpoint.

\subsection{Selection and Profiling Protocol}
\vspace{-1mm}
All calibrators in Table~\ref{tab:calibrator_comparison} of the main paper share 238 test initializations (1 May--24 December 2018), five variables, and four leads.
Development initializations end on 22 April; 23--30 April forms an eight-day unscored embargo.
Training-time networks follow the offline protocol in Appendix~\ref{app:calibration_baselines}.
Each online adapter selects a learning rate from $\{10^{-5},10^{-4},10^{-3}\}$ using pooled 20-cell macro nRMSE on 41 validation initializations (13 March--22 April).
Their four-lead labels are complete by 30 April, before the first test issuance.
Selected model and Adam states, together with COSA~\cite{im2026cosa} issuance contexts, continue through the embargo and testing, with one update per fully verified four-lead window.
TAFAS~\cite{kim2025tafas} uses $10^{-4}$ on both backbones; COSA and PETSA~\cite{medeiros2025petsa} use $10^{-3}$ on both.
ST-TTC~\cite{chen2025stttc} uses $10^{-4}$ on GraphCast~\cite{lam2023learning} and $10^{-3}$ on Pangu~\cite{bi2023accurate}.
Ordinary grid-point MOS~\cite{glahn1972mos} selects its ridge coefficient from $\{0,10^{-5},10^{-4},10^{-3},0.01,0.05,0.1,1\}$ on the same validation set; both backbones select zero.
Neural inputs use fixed 1979--2016 means and standard deviations; diagnostic ACC uses the WeatherBench~2 1959--2015 daily climatology.

Calibration-only profiling uses one NVIDIA L4, FP32 with TF32 disabled, and the common $4\times5\times121\times240$ workload.
Each measurement follows 10 warm-up iterations and uses 30 measured repetitions.
Online costs include one fixed Adam step at $10^{-4}$, independently of accuracy selection.
SphereTTC updates the incoming coefficients, fits and gates 64 verified pairs reconstructed from complete daily ERA5 truth, and applies the correction.
The timed SphereTTC output matches the reference implementation bit for bit.

\paragraph{Matched feedback availability.}
A target-day mean comprises the four ERA5 fields at 00, 06, 12, and 18 UTC and becomes eligible at 00 UTC the following day. For example, a forecast issued at 00 UTC on 1 May has a day-one target on 2 May, available at 00 UTC on 3 May, and a day-seven target on 8 May, available on 9 May. A per-lead history admits the day-one pair on 3 May. A complete-window history admits both pairs on 9 May.

\begin{center}
\begin{minipage}{\textwidth}
\captionof{table}{Feedback availability and update schedule. The issue time is $s$ and the lead is $\ell$ days. The rightmost column counts gradient steps during testing.}
\label{tab:app_feedback_schedule}
\small\setlength{\tabcolsep}{4pt}\renewcommand{\arraystretch}{1.13}
\begin{tabularx}{\textwidth}{@{}p{117pt}XXc@{}}
\toprule
Method & First eligible verification & Test-time admission or update & Steps\\
\midrule
Raw & No feedback used & Unmodified forecast & 0\\
SphereTTC (per lead) & $s+\ell+1$ days, separately by lead & Admit each mature pair to its lead's 64-pair history & 0\\
SphereTTC (complete window) & $s+8$ days, common to all leads & Admit all four pairs to their 64-pair histories together & 0\\
Running-mean bias, MOS & $s+\ell+1$ days, separately by lead & Admit each mature pair to its lead's 64-pair history & 0\\
TAFAS, ST-TTC, COSA, PETSA & $s+8$ days, common to all leads & Update once for each newly complete source window & 1\\
DRN, BC-UNet, PnP-Corrector, ASTAFN & Verified labels used during pre-test fitting & Frozen during testing & 0\\
\bottomrule
\end{tabularx}
\end{minipage}
\end{center}

To assess sensitivity to feedback timing, we replay both SphereTTC schedules on the same 238 initializations, two backbones, five variables, and four leads.
Bandwidth, blend cap, 64-pair memory, decay, and gate settings remain fixed in this NVIDIA L4 Colab replay.
Forecast fields, targets, dates, and climatology match the archived COSA output-adapter run, whose daily sufficient statistics we reuse.
COSA uses a learning rate of $10^{-3}$ on both backbones and 350 complete-window updates over the stream, including warm-up.
Its archived nRMSE reductions are 1.25\% for GraphCast and 1.88\% for Pangu; the corresponding main-paper Table~\ref{tab:calibrator_comparison} values are 1.15\% and 1.58\%.
Table~\ref{tab:app_matched_feedback} uses the paired archive scores throughout.

\begin{center}
\begin{minipage}{\textwidth}
\captionof{table}{Feedback-matched comparison on 238 initializations per backbone. Absolute scores pool daily statistics before averaging across variables and, for the overall row, leads. Differences $\Delta$ are complete-window SphereTTC minus archived COSA, scaled by $10^4$, with pointwise 95\% paired 14-day circular block-bootstrap intervals (2,000 replicates). $\Delta$ nRMSE$<0$ and $\Delta$ ACC$>0$ favor SphereTTC.}
\label{tab:app_matched_feedback}
\centering\fontsize{8.5}{10}\selectfont\setlength{\tabcolsep}{2pt}\renewcommand{\arraystretch}{1.10}
\begin{tabular*}{\textwidth}{@{\extracolsep{\fill}}llrrrrrrll@{}}
\toprule
& & \multicolumn{2}{c}{SphereTTC (per lead)} & \multicolumn{2}{c}{SphereTTC (complete)} & \multicolumn{2}{c}{COSA} & \multicolumn{2}{c}{Paired difference $\times 10^4$}\\
\cmidrule(lr){3-4}\cmidrule(lr){5-6}\cmidrule(lr){7-8}\cmidrule(l){9-10}
Backbone & Day & nRMSE & ACC & nRMSE & ACC & nRMSE & ACC & $\Delta$ nRMSE [95\% CI] & $\Delta$ ACC [95\% CI]\\
\midrule
GraphCast & 1 & 0.03771 & 0.99252 & 0.03772 & 0.99251 & 0.03784 & 0.99241 & $-1.20\,[-1.54,-0.89]$ & $+0.98\,[+0.77,+1.21]$\\
 & 3 & 0.08642 & 0.96781 & 0.08642 & 0.96781 & 0.08650 & 0.96767 & $-0.85\,[-1.31,-0.39]$ & $+1.42\,[+0.68,+2.21]$\\
 & 5 & 0.16106 & 0.89301 & 0.16108 & 0.89298 & 0.16031 & 0.89236 & $+7.65\,[+4.43,+11.29]$ & $+6.21\,[+1.40,+10.45]$\\
 & 7 & 0.24089 & 0.74797 & 0.24089 & 0.74797 & 0.24113 & 0.74647 & $-2.38\,[-18.35,+12.17]$ & $+14.93\,[-4.54,+32.37]$\\
 & \textbf{Overall} & 0.13152 & 0.90033 & 0.13153 & 0.90032 & 0.13145 & 0.89973 & $+0.80\,[-3.73,+4.90]$ & $+5.88\,[-0.03,+11.34]$\\
\midrule
Pangu & 1 & 0.04220 & 0.99017 & 0.04224 & 0.99013 & 0.04275 & 0.98976 & $-5.09\,[-5.50,-4.70]$ & $+3.66\,[+3.29,+4.05]$\\
 & 3 & 0.09443 & 0.96122 & 0.09448 & 0.96113 & 0.09481 & 0.95975 & $-3.35\,[-4.42,-2.21]$ & $+13.82\,[+11.93,+15.87]$\\
 & 5 & 0.17371 & 0.87643 & 0.17376 & 0.87631 & 0.17200 & 0.87387 & $+17.59\,[+12.19,+22.85]$ & $+24.40\,[+15.06,+32.51]$\\
 & 7 & 0.25652 & 0.71743 & 0.25652 & 0.71743 & 0.25652 & 0.71823 & $-0.02\,[-21.79,+20.38]$ & $-7.94\,[-33.12,+16.03]$\\
 & \textbf{Overall} & 0.14172 & 0.88631 & 0.14175 & 0.88625 & 0.14152 & 0.88540 & $+2.28\,[-4.25,+8.10]$ & $+8.48\,[-0.25,+16.44]$\\

\bottomrule
\end{tabular*}
\end{minipage}
\end{center}

Aggregate SphereTTC scores change little when all leads wait for complete verification.
The maximum changes across the two backbones are $0.000034$ nRMSE and $0.000065$ ACC.
Under this shared availability rule, SphereTTC retains lower nRMSE and higher ACC than COSA at days one and three, with pointwise intervals excluding zero.
At day five, COSA gives lower nRMSE and SphereTTC higher ACC; the two SphereTTC schedules coincide at day seven.
Across all leads, COSA has lower nRMSE point estimates and SphereTTC higher ACC point estimates on both backbones.
The overall intervals include zero for both metrics and both backbones.

\Needspace*{9\baselineskip}
\section{Spatial and Spectral Diagnostics}
\label{app:spatial_spectral}
\vspace{-1mm}
Spatial maps and error spectra complement aggregate scores by resolving geographic and scale-dependent error patterns~\cite{han2023steerer}.
The main field comparison uses GraphCast~\cite{lam2023learning} and Pangu~\cite{bi2023accurate} forecasts initialized on 30 May 2018 at a one-day lead, with 64 eligible history pairs.
Its MSE agrees with the reference evaluator within 0.001\%.
Figure~\ref{fig:spatial_rmse} summarizes spatial RMSE and zonal profiles over the diagnostic cohort; Figure~\ref{fig:spectral_diagnostics} shows the five-day error spectra.
\par\vspace{3pt}\noindent\begin{minipage}{\textwidth}
\centering
\includegraphics[width=\textwidth]{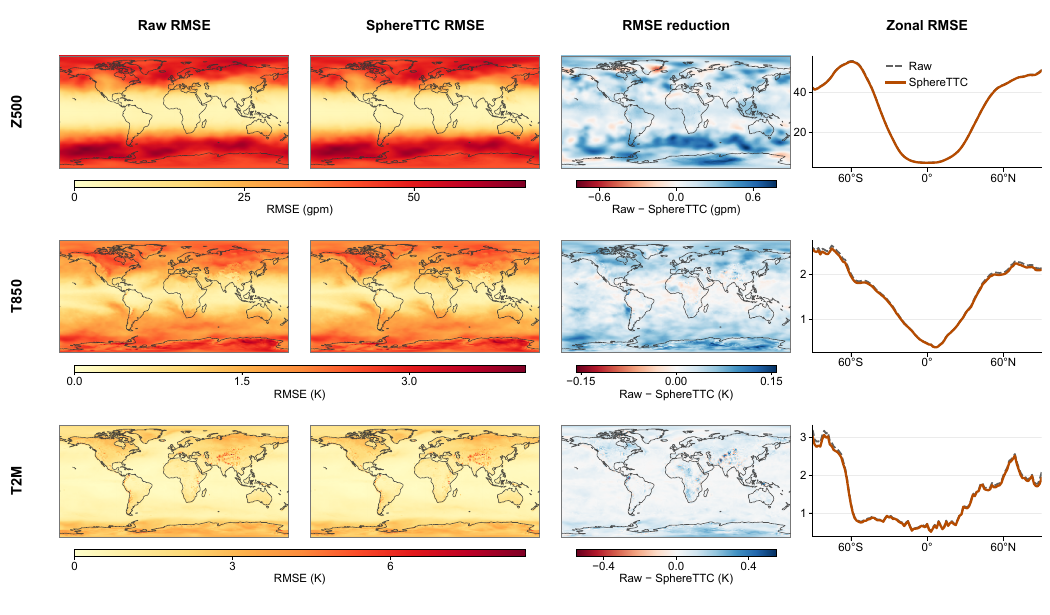}
\captionof{figure}{Spatial RMSE and zonal profiles for 299 GraphCast~\cite{lam2023learning} dates and four leads. The difference maps show Raw minus SphereTTC.}
\label{fig:spatial_rmse}
\end{minipage}\par\vspace{3pt}
\vspace{-2pt}
\par\vspace{3pt}\noindent\begin{minipage}{\textwidth}
\centering
\includegraphics[width=0.93\textwidth]{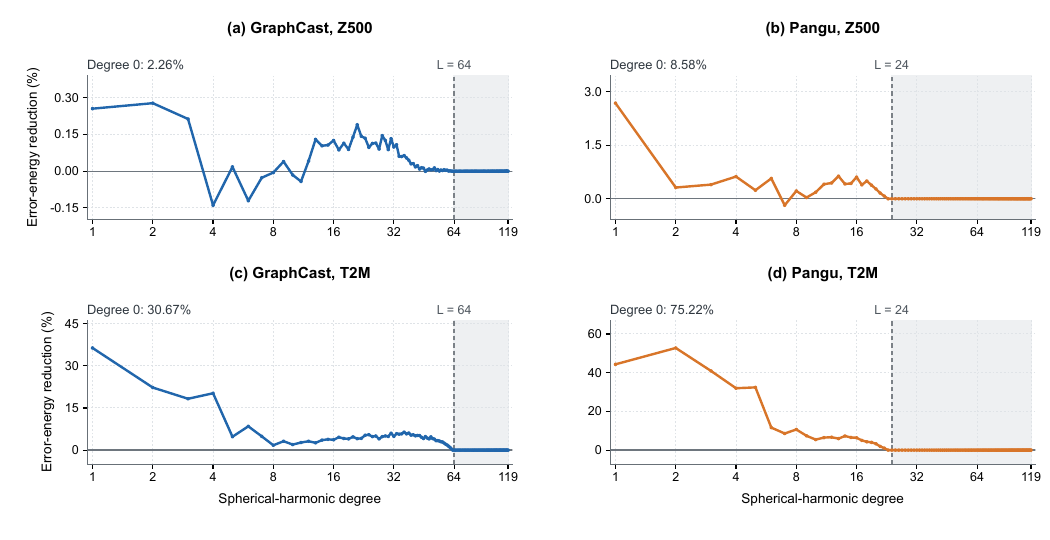}
\captionof{figure}{Five-day error spectra on the 287-date diagnostic cohort. The restricted correction changes low degrees while the high-degree tail remains essentially unchanged.}
\label{fig:spectral_diagnostics}
\end{minipage}\par\vspace{3pt}

\Needspace*{9\baselineskip}
\section{Verification Gating}
\label{app:gate_audit}
\vspace{-1mm}
We evaluate verification gating through the distribution of per-outcome RMSE changes on the matched diagnostic cohort.
\par\vspace{3pt}\noindent\begin{minipage}{\textwidth}
\centering
\includegraphics[width=0.80\textwidth]{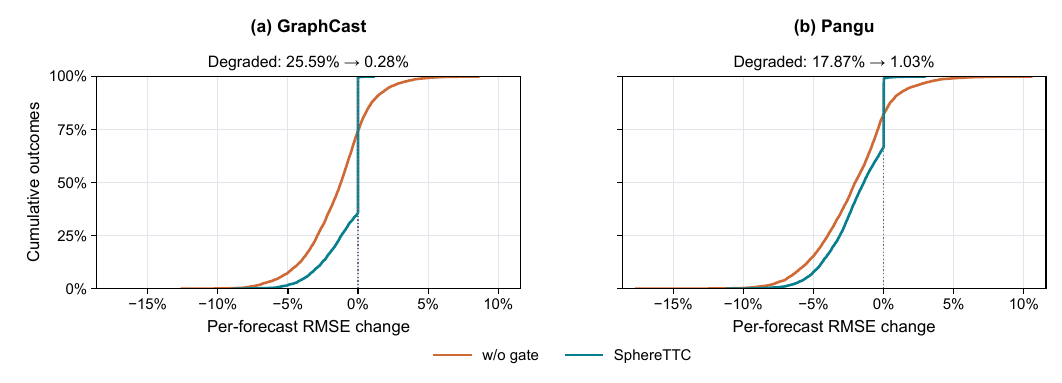}
\captionof{figure}{Per-outcome RMSE-change distributions on the matched diagnostic cohort. Gating suppresses the worsening tail while retaining beneficial corrections.}
\label{fig:app_gate_distribution}
\end{minipage}\par\vspace{3pt}
\noindent Gating reduces the fraction of degraded outcomes from 25.59\% to 0.28\% for GraphCast~\cite{lam2023learning} and from 17.87\% to 1.03\% for Pangu~\cite{bi2023accurate}.
This reliability gain accompanies higher ACC and smaller aggregate nRMSE reductions than ungated calibration.

\paragraph{Issuance-time replay.}
The current gate scores affine proposals in harmonic coefficient space on the most recent holdout of the history available at the current issuance. For the strict replay, each historical forecast issued at $u$ receives an affine proposal fitted on the latest $K$ pairs that had completed verification by $u$. We store the proposal after applying the taper and inverse transform and score it with latitude-weighted grid MSE once its target day completes. At the current issuance $t$, the gate uses only stored scores completed by $t$, taking the newest eligible scores up to the current holdout size. Thus the historical fits follow their own issuance cutoffs and the gate scores the spatial correction used at deployment.

We compare four variants on the 287 common origins from 13 March to 24 December 2018 (5,740 outcomes per backbone). All use the same correction fits and taper at test issuance. The ungated control uses the maximum blend. The fixed scalar is the mean gate strength among eligible issuance--lead pairs before 13 March, pooled across the five variables for each backbone. It is 0.0496 for GraphCast and 0.3224 for Pangu. The mean test gate strengths for the current method are 0.1060 and 0.3360, respectively.

\begin{center}
\begin{minipage}{\textwidth}
\captionof{table}{Gate replay on 287 matched initializations per backbone. Corrected denotes a nonzero grid correction. Degradation denotes higher latitude-weighted RMSE than Raw. The last column gives the mean relative RMSE increase among degraded outcomes.}
\label{tab:app_gate_replay}
\centering\small\setlength{\tabcolsep}{3pt}\renewcommand{\arraystretch}{1.13}
\begin{tabular*}{\textwidth}{@{\extracolsep{\fill}}llrrrrrr@{}}
\toprule
Backbone & Blend rule & nRMSE & ACC & \shortstack{Corrected\\(\%)} &
\shortstack{Degraded\\(\% of all)} & \shortstack{Degraded\\(\% of corrected)} &
\shortstack{Mean increase\\if degraded (\%)}\\
\midrule
GraphCast & No gate & 0.13124 & 0.89994 & 100.00 & 25.59 & 25.59 & 1.37\\
 & Fixed blend & 0.13342 & 0.89984 & 100.00 & 3.08 & 3.08 & 0.04\\
 & Current gate & 0.13229 & 0.90059 & 36.06 & 0.28 & 0.77 & 0.39\\
 & Strict replay & 0.13188 & 0.90090 & 60.05 & 0.75 & 1.25 & 0.42\\
\midrule
Pangu & No gate & 0.14153 & 0.88479 & 100.00 & 17.87 & 17.87 & 1.54\\
 & Fixed blend & 0.14242 & 0.88715 & 100.00 & 3.43 & 3.43 & 0.38\\
 & Current gate & 0.14223 & 0.88716 & 67.63 & 1.03 & 1.52 & 0.55\\
 & Strict replay & 0.14169 & 0.88701 & 86.29 & 2.46 & 2.85 & 0.77\\
\bottomrule
\end{tabular*}
\end{minipage}
\end{center}

The current gate has the lowest degradation frequency in both the full and corrected subsets.
On Pangu, the fixed and current mean blend strengths are close, yet the current gate lowers degradation among corrected outcomes from 3.43\% to 1.52\% and improves nRMSE.
Selective correction is accompanied by lower degradation; the fixed-blend comparison combines selection and strength, particularly on GraphCast where mean strengths differ more.
Strict replay improves nRMSE on both backbones and applies corrections more often, with a higher degradation rate than the current gate.
It changes both historical fitting cutoffs and the proposal-scoring space.
Both gates reduce degradation frequency and mean severity relative to ungated calibration.

\clearpage
\section{Matched Spectral Control}
\label{app:matched_spectral_control}
\vspace{-1mm}
The matched spectral control tests the contribution of the fitted affine scale.
The additive control fixes this scale to one while retaining the same bandwidth, 64 eligible pairs, weights, chronological gate, and taper.
Both methods use 238 dates (1 May--24 December 2018), five variables, and four leads, with verification admitted after target-day completion and no test-set tuning.
Each backbone contributes 4,760 initialization--variable--lead outcomes.
Degradation denotes increased latitude-weighted MSE relative to Raw, equivalently an increase in RMSE.

\begin{center}
\begin{minipage}{\textwidth}
\centering
\captionsetup{skip=3pt}
\captionof{table}{Matched spectral control. Relative nRMSE and ACC gains are higher-is-better; degraded-outcome frequency is lower-is-better.}
\label{tab:matched_spectral_control}
\small
\setlength{\tabcolsep}{2pt}
\begin{tabularx}{\textwidth}{lCCC}
\toprule
Correction & $\Delta$nRMSE & $\Delta$ACC & Degraded\\
& (\%) & (\%) & (\%)\\
\midrule
\multicolumn{4}{l}{GraphCast~\cite{lam2023learning}}\\
Gated SH additive & 0.110 & 0.053 & 1.13\\
SphereTTC & 1.194 & 0.125 & 0.25\\
\midrule
\multicolumn{4}{l}{Pangu~\cite{bi2023accurate}}\\
Gated SH additive & 0.466 & 0.258 & 1.62\\
SphereTTC & 1.747 & 0.173 & 1.09\\
\bottomrule
\end{tabularx}
\end{minipage}
\end{center}

\begin{center}
\begin{minipage}{\textwidth}
\captionof{table}{Paired differences between affine and additive spectral calibration. Positive values favor the affine model; intervals are pointwise paired 95\% intervals in percentage points.}
\label{tab:matched_spectral_ci}
\small\setlength{\tabcolsep}{5pt}
\begin{tabular*}{\textwidth}{@{\extracolsep{\fill}}llcc@{}}
\toprule
Backbone & Improvement metric & Full minus additive (pp) & 95\% interval\\
\midrule
GraphCast~\cite{lam2023learning} & $\Delta$nRMSE & +1.084 & [+0.915, +1.257]\\
GraphCast~\cite{lam2023learning} & $\Delta$ACC & +0.072 & [+0.053, +0.091]\\
Pangu~\cite{bi2023accurate} & $\Delta$nRMSE & +1.281 & [+1.111, +1.466]\\
Pangu~\cite{bi2023accurate} & $\Delta$ACC & -0.085 & [-0.118, -0.051]\\
\bottomrule
\end{tabular*}
\end{minipage}
\end{center}

We use 2,000 paired circular block-bootstrap replicates of 14 dates, with seed 20261031.
Methods, channels, and leads share each sampled sequence; error and anomaly statistics are pooled before macro aggregation.
The affine correction gives larger nRMSE gains and fewer degraded outcomes on both backbones; the paired intervals support its nRMSE advantage.
Its ACC gain is also higher on GraphCast, whereas the additive correction gives the higher ACC gain on Pangu~\cite{bi2023accurate}.
Fitting the affine scale therefore contributes beyond an additive spectral adjustment, with the clearest benefit in nRMSE.

\section{Calibration Scope}
\label{app:limitations}
\vspace{-1mm}
SphereTTC adjusts each scalar channel within the selected harmonic bandwidth and preserves the frozen forecaster's finer-scale detail. Updates follow the admission rule specified for each evaluation cohort. The reported scores assess deterministic forecast skill at the daily time scale across the evaluated backbones, leads, and feedback schedules.
\endgroup

\end{document}